\documentclass[lettersize,journal]{IEEEtran}
\usepackage{amsmath,amsfonts}
\usepackage{algorithmic}
\usepackage{algorithm}
\usepackage{array}
\usepackage[caption=false,font=normalsize,labelfont=sf,textfont=sf]{subfig}
\usepackage{textcomp}
\usepackage{stfloats}
\usepackage{url}
\usepackage{verbatim}
\usepackage{graphicx}
\usepackage{cite}
\usepackage{diagbox}
\usepackage{threeparttable}

\begin{document}

\title{Adaptive Denoising-Enhanced LiDAR Odometry for Degeneration Resilience in Diverse Terrains}

\author{Mazeyu Ji, Wenbo Shi, Yujie Cui, Chengju Liu and Qijun Chen,~\IEEEmembership{Senior Member,~IEEE}

\thanks{This paper was supported by the National Natural Science Foundation of China under Grant 62233013, Grant 62073245, and Grant 62173248; in part by the Shanghai Science and Technology Innovation Action Plan under Grant 22511104900; in part by the Shanghai Municipal Science and Technology Major Project under Grant 2021SHZDZX0100; in part by the Fundamental Research Funds for the Central Universities; and in part by the Research and Development Center of Transport Industry of New Generation of Artificial Intelligence Technology under Grant 202201H. (Mazeyu Ji and Wenbo Shi contributed equally to this work.) (Corresponding authors: Chengju Liu; Qijun Chen.)}
\thanks{Mazeyu Ji, Wenbo Shi, Yujie Cui, and Qijun Chen are with the School of Electronics and Information Engineering, Tongji University, Shanghai 201804, China (e-mail: jimazeyu0@gmail.com; 1810842@tongji.edu.cn; jshmcyj@tongji.edu.cn; qjchen@tongji.edu.cn). }

\thanks{Chengju Liu is with the School of Electronics and Information Engineering, Tongji University, Shanghai 201804, China, and also with the Tongji Research Institute of Artificial Intelligence (Suzhou), Jiangsu 215131, China (e-mail: liuchengju@tongji.edu.cn). }

}


\maketitle

\begin{abstract}

The flexibility of Simultaneous Localization and Mapping (SLAM) algorithms in various environments has consistently been a significant challenge. To address the issue of LiDAR odometry drift in high-noise settings, integrating clustering methods to filter out unstable features has become an effective module of SLAM frameworks. However, reducing the amount of point cloud data can lead to potential loss of information and possible degeneration. As a result, this research proposes a LiDAR odometry that can dynamically assess the point cloud's reliability. The algorithm aims to improve adaptability in diverse settings by selecting important feature points with sensitivity to the level of environmental degeneration. Firstly, a fast adaptive Euclidean clustering algorithm based on range image is proposed, which, combined with depth clustering, extracts the primary structural points of the environment defined as ambient skeleton points.  Then, the environmental degeneration level is computed through the dense normal features of the skeleton points, and the point cloud cleaning is dynamically adjusted accordingly. The algorithm is validated on the KITTI benchmark and real environments, demonstrating higher accuracy and robustness in different environments.

\end{abstract}


\section{Introduction}
\IEEEPARstart{S}{LAM} (Simultaneous Localization and Mapping) technology enables real-time environment modeling and autonomous localization, providing crucial perception and decision-making capabilities to robots. It holds significant importance in various fields such as robot navigation, 3D modeling, and augmented reality. With advancements in technology and hardware development, 3D LiDAR SLAM has become the mainstream choice.

In the past decade, numerous outstanding 3D LiDAR SLAM algorithms\cite{loam,loam-livox,suma,suma++,lio-sam,fast-lio,fast-lio2} have emerged. They mainly focus on how to accurately match point clouds and how to globally optimize the pose. These algorithms effectively improve the accuracy of localization and mapping, but most of them do not prioritize adaptability to the environment, often requiring repeated parameter adjustments. In practical applications, SLAM algorithms need to perform localization and mapping in diverse environments, such as urban areas, forests, tunnels, and highways. Each environment has its unique terrain, structure, and sensor reflection characteristics, all of which impact the performance of SLAM algorithms. When facing the challenges posed by variable terrains, LeGO-LOAM\cite{lego-loam} has demonstrated remarkable capabilities. This algorithm leverages segmentation techniques to extract reliable point clouds, thereby enhancing the accuracy of localization and mapping. The concept introduced by LeGO-LOAM has also been incorporated into some subsequent SLAM frameworks\cite{imls-slam,t-loam,pago-loam}. The cleaning operation applied to the point cloud effectively enhances the algorithm's adaptability to the challenges posed by complex environments.

\begin{figure}[t!]
\centering
\subfloat[]{\includegraphics[width=1.5in]{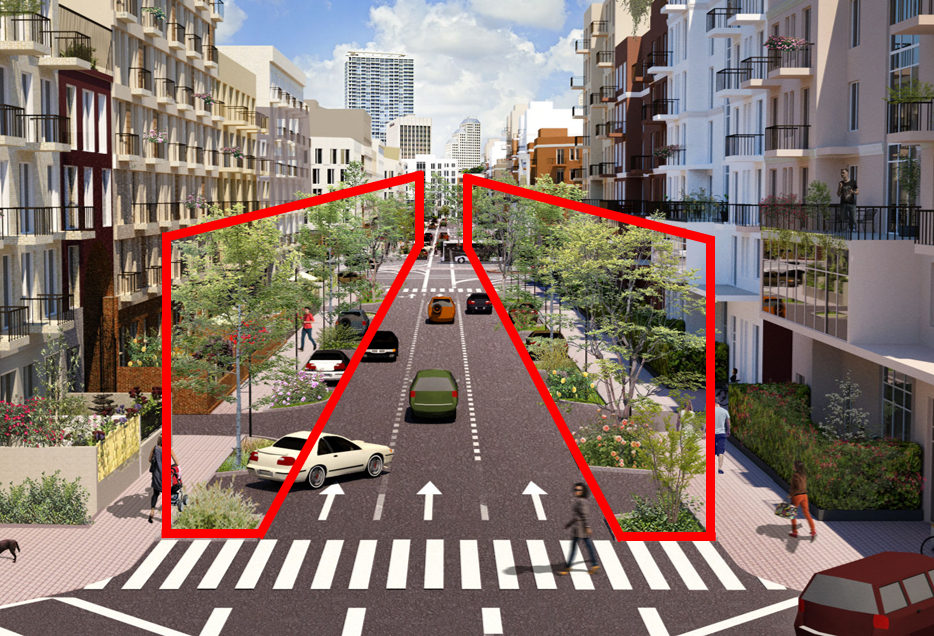}%
\label{fig_first_case}}
\hfil
\subfloat[]{\includegraphics[width=1.5in]{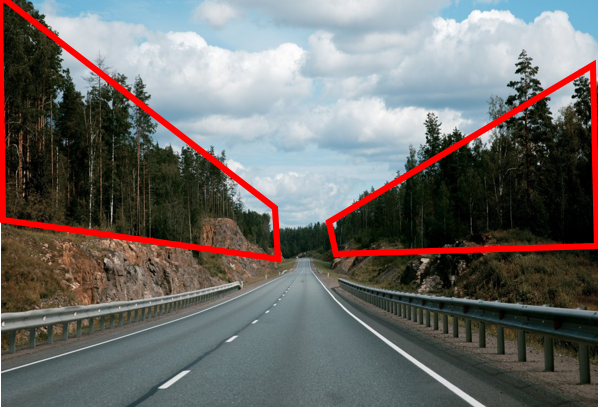}%
\label{fig_second_case}}
\caption{The importance of the same object varies a lot in different scenes. In the scene (a), there are already sufficient features provided, and the leaves introduce matching noise. In the scene (b), if the features provided by the leaves are not fully utilized, the scene will be too simple and leads to degeneration.}
\label{fig_intro}
\end{figure}

However, it is inappropriate to use a fixed and unchanging point cloud cleaning criterion because the reliability of the same features often varies greatly in different scenes. As shown in Fig. \ref{fig_intro}, leaves, as unreliable features, can introduce matching noise in most complex scenes, but they become essential references for matching in others, like tunnels and highways, which can be defined as degeneration scenes. Degeneration refers to the reduction in the quantity and availability of useful point cloud information, which could lead to constraint deficiency and characterization insufficiency. The excessive cleaning of point clouds can exacerbate the degeneration of these scenes, as a result, it is essential to find the dynamic balance between information loss and reliable features. Some work\cite{Degenerate1,Degenerate2} now take into account the dynamic weighting of point cloud features, but they are based solely on local point cloud structure without considering whether the point cloud as a whole has experienced degeneration. Therefore, detecting degeneration scene becomes crucial. Several related studies have proposed solutions for detecting degeneration scenes\cite{lu-idea,bb-hessian,censi-fisher,ZS,loam,line-feature1,line-feature2,line-feature3,line-feature4,cylinder,dncsm,adalio}. Among them, methods based on point cloud matching\cite{lu-idea,bb-hessian,censi-fisher,ZS,loam} rely on the accuracy of point cloud matching and cannot judge the degree of degeneration in the preprocessing stage. On the other hand, feature-based scene degeneration detection\cite{line-feature1,line-feature2,line-feature3,line-feature4,cylinder,dncsm,adalio} faces challenges such as time-consuming feature extraction and high noise in 3D point clouds. Consequently, investigating how to rapidly extract reliable features from point clouds for analyzing the degree of scene degeneration is a worthwhile research problem.

Considering the aforementioned issues, we propose a novel adaptive point cloud denoising-enhanaced algorithm for LiDAR odometry with degeneration resilience, which efficiently extracts the main objects from the scene point cloud, and combines the structural features to dynamically adjust the SLAM algorithm's point cloud filtering process according to the degeneration level. Different from many other LiDAR odometry algorithms, we provide a unique way of environmental understanding, which allows for seamless integration with various SLAM systems. We incorporated this module into the LeGO-LOAM algorithm framework and evaluated the algorithm's performance using both the KITTI benchmark\cite{kitti} and real-world scenarios. The results showed improved accuracy and robustness in comparison to the original algorithm. Specifically, the main contributions of this paper are as follows:

\begin{enumerate}
\item{We have noticed the orderliness of the LiDAR point cloud. By combining the point cloud sequence and physical characteristics, we have implemented a fast Euclidean clustering based on range images. This approach ensures real-time performance while increasing the accuracy of point cloud segmentation.}
\item{We have observed that the overall information about the environment is embedded within the structural skeleton of the point cloud. By combining clustering information, we can rapidly extract the main environmental structures, providing a foundation for the analysis of point cloud information.}
\item{We extend the normal vector based degeneration scene detection method to the point cloud skeleton analysis, dynamically adjusting the point cloud pre-segmentation strategy by calculating the degree of scene degeneration.}
\end{enumerate}

The subsequent sections of this paper are structured as follows. Section II presents a review of the related work. In Section III, the overall framework of the algorithm is outlined. Detailed elaboration of the proposed approach is provided in Sections IV and V, explaining the entire process. Section VI presents a diverse set of experiments conducted to analyze the results. Finally, Section VII concludes the article.

\section{RELATED WORK}
\subsection{3D LiDAR Odometry}
LiDAR odometry is a fundamental component of 3D SLAM systems, and a robust and efficient LiDAR odometry serves as the foundation for such systems. 

In recent years, numerous outstanding LiDAR odometry methods\cite{loam,lego-loam,loam-livox,suma,suma++} have been proposed in the field. Among them, the mainstream 3D LiDAR odometry approaches are primarily based on LiDAR odometry and mapping(LOAM)\cite{loam} introduced by Zhang et al. LOAM extracts sparse corner and planar features based on the smoothness of points. It then employs point-to-line and point-to-plane tactics for scan matching. Afterwards, LOAM-Livox\cite{loam-livox} extends LOAM to LiDARs with small Field of View (FoV) and irregular samplings, resulting in improved accuracy and efficiency. LeGO-LOAM\cite{lego-loam} builds upon the foundation of LOAM by introducing ground constraints, significantly improving efficiency, and reducing drift along the z-axis. The LOAM-based LiDAR odometry has also been subsequently applied to some SLAM systems that incorporate 3D LiDARs\cite{lio-sam,fast-lio,fast-lio2,r2live,r3live}. Apart from the LOAM-based LiDAR odometry, the most representative approach is semantic-based\cite{suma,suma++}, which leverages semantic information to better comprehend the environment and mitigate dynamic obstacles. Nevertheless, its high computing cost reduces the practicality of the algorithm.

However, SLAM in different terrain environments remains a challenging problem. Many algorithms have been developed for specific environments\cite{dare-slam,forest-slam,mine-slam,indoor-slam}, but they often require a significant number of strict parameters and constraints tailored to the specific characteristics of the environment. Moreover, the same algorithm often requires frequent parameter adjustments when applied in different environments. To address the challenge, LeGO-LOAM introduced point cloud segmentation into 3D LiDAR SLAM. During point cloud preprocessing, it utilizes segmentation techniques to remove unstable small feature clusters, reducing the impact of noise and errors. This approach enhances algorithm robustness and allows it to perform well in various terrain environments. Due to its ability to aid in understanding and modeling complex scenes, point cloud segmentation has also been applied in many subsequent SLAM frameworks\cite{imls-slam,t-loam,pago-loam}. The point cloud filtering operation effectively improves these algorithms' adaptability to environmental complexity.

However, removing points from the point cloud inevitably leads to information loss. In feature-scarce degeneration environments, preserving more reliable features is essential to fully leverage the point cloud information. To address this, we introduce a degeneration scene detection module into 3D LiDAR Odometry. This module dynamically adjusts the point cloud removal level based on the degeneration degree, maximizing the utilization of point cloud information, and improving the overall system robustness.

\begin{figure*}[t!]
\centering
\includegraphics[width=6.0in]{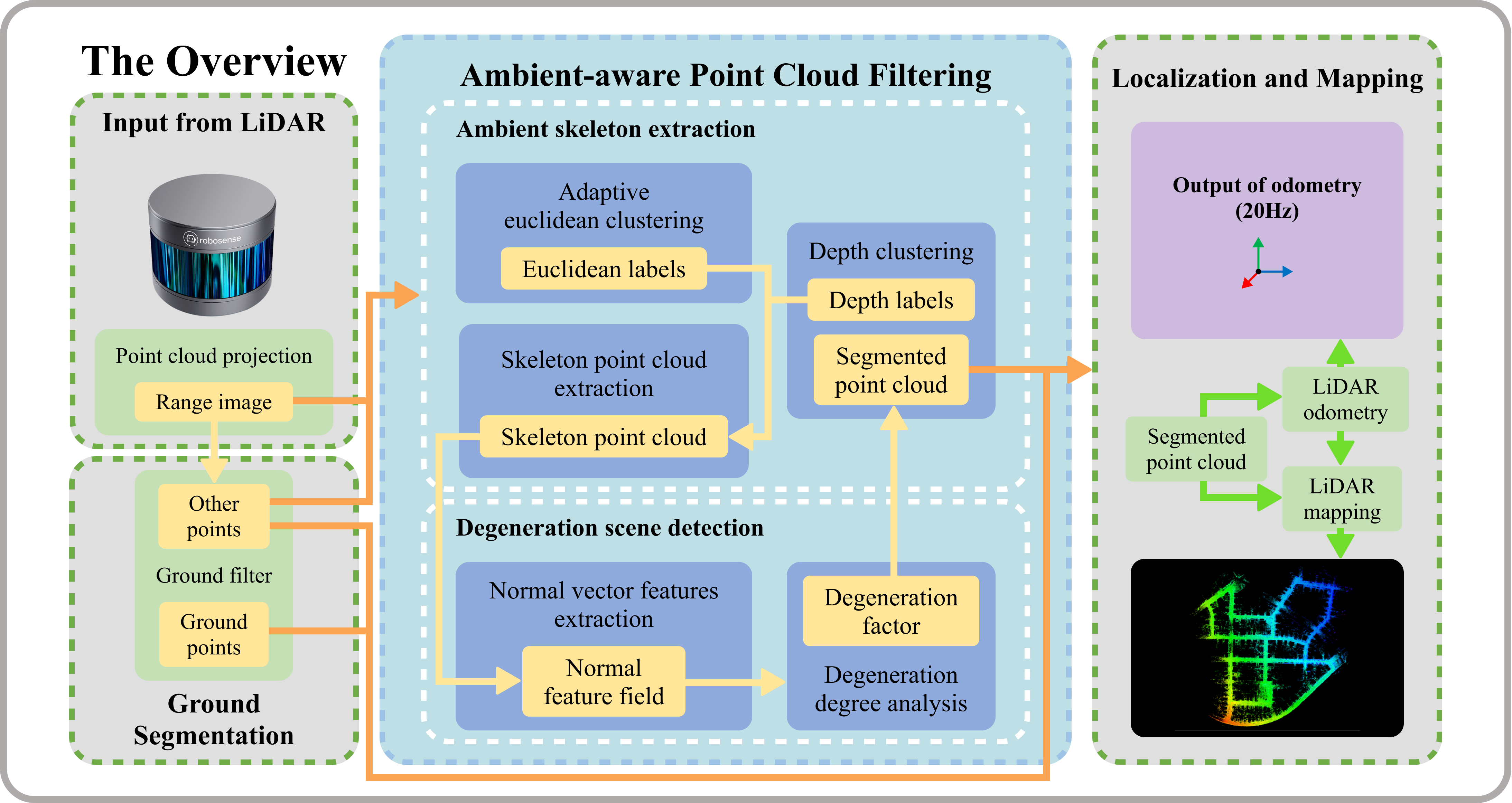}
\caption{We present the overall framework of our algorithm, focusing on the main contribution of ambient-aware point cloud filtering, which is integrated into the SLAM framework of LeGO-LOAM. The ambient-aware filtering consists of two key components: ambient skeleton extraction and degeneration scene detection. The former extracts point cloud data that represents the fundamental structure of the scene, while the latter analyzes the degree of scene degeneration and dynamically adjusts the threshold for point cloud segmentation and denoising. Our algorithm provides the dynamically denoised point cloud to the LiDAR odometry module, resulting in more accurate and robust localization performance.}
\label{overview}
\end{figure*}

\subsection{Point Cloud Segmentation}

Point Cloud Segmentation is a crucial step in 3D laser point cloud processing. It is widely used in tasks like object detection, pose estimation, motion analysis, and 3D reconstruction. In SLAM, it always helps filter out unstable objects for noise reduction while maintaining real-time performance and high object segmentation capabilities.

Currently, point cloud segmentation algorithms fall into two categories: deep learning based and traditional methods. Although numerous deep learning approaches\cite{squeezeseg,voxelnet,pointnet,rangenet} continuously improve segmentation accuracy, their high computational complexity makes them unsuitable for SLAM applications, which demand high computational efficiency. On the other hand, traditional point cloud segmentation methods have undergone extensive development and have yielded many efficient algorithms that are more commonly used in processing LiDAR data.

Traditional point cloud segmentation can be categorized into two main types: spatial-based clustering and connected component labeling (CCL) based clustering. The spatial-based clustering methods directly utilize the 3D spatial coordinates of the points. These methods often employ algorithms based on spatial distance and density\cite{Euclidean,autoware,flic,adaptive,dbscan, super-voxel,cvc}. However, these algorithms typically require extensive point cloud position queries, resulting in slow processing speeds. On the other hand, the CCL-based clustering methods\cite{flic,run,depth,fspb} map the point cloud to a 2D range image and apply the principles of connected component labeling for data processing. During the point cloud scanning process, the neighboring relationships are established based on the physical characteristics of the LiDAR, creating a form of weak adjacency. Despite this, the advantage lies in their fast query speed, making them computationally efficient.

Numerous researchers have compared various clustering algorithms regarding their segmentation capabilities and found that Euclidean-based clustering algorithms often exhibit better performance in segmentation tasks\cite{eu-better,seg-bench,adaptive}. However, they are constrained by computational speed and sensitivity to distance. Inspired by the work of\cite{flic,fspb} which utilized CCL weak adjacency for accelerated queries, and the work\cite{adaptive} that dynamically adjusted thresholds, this paper proposes a fast adaptive Euclidean clustering algorithm based on range images. This algorithm achieves high object segmentation accuracy while ensuring real-time performance.

\subsection{Degeneration Scene Detection}

Degeneration scenes are an inevitable challenge in SLAM. When a robot moves and the point cloud's pose constraints are insufficient, it results in significant pose uncertainty.  Effectively detecting and handling degeneration scenes is crucial to address this issue effectively.

The detection of degeneration scenes can be broadly classified into two methods. The first method is based on aligning point clouds between two frames. This approach follows Lu's idea of equating scene degeneracy to uncertainty in point cloud matching\cite{lu-idea}. Bengtsson and Baerveldt et al.\cite{bb-hessian} utilized the Hessian of the matching covariance matrix with the iterative dual correspondence (IDC) approach, while Censi\cite{censi-fisher} used the Fisher information matrix with the iterative closest point (ICP) method. Zhang and Singh\cite{ZS} employed eigenvalues to design a fused state uncertainty estimation approach based on LOAM\cite{loam}. However, all these algorithms require aligning point cloud scans between two frames, introducing errors from the matching process and leading to higher uncertainty, which affects degeneration scene detection. In this paper, to optimize the point cloud matching process and reduce computational load, we propose to assess scene degeneracy during the preprocessing stage rather than using a two-step point cloud matching approach.

Another mainstream approach for degeneration scene detection is based on point cloud geometric features. Line features are extensively utilized\cite{line-feature1,line-feature2,line-feature3,line-feature4} and exhibit excellent performance in structured environments. Nevertheless, in unstructured scenes or with noisy LiDAR point clouds, line features may not fit smoothly, resulting in a notable impact on degeneration detection. Tolga et al.\cite{cylinder} proposed high-order features like cylinder features, but the computations are too complex. Shi et al.\cite{dncsm} used the principal component analysis(PCA) of normal vector features to assess the degree and direction of scene degeneracy, providing a simple and efficient method for 2D degeneration scene detection. AdaLIO\cite{adalio} applied a similar idea in 3D SLAM, but it used radius search to find neighboring points and compute normal vectors, leading to significant computing cost in 3D. Additionally, in 3D SLAM, the problems are often more complex with more noise in point clouds, resulting in highly imbalanced distributions of normal vectors\cite{imba1,imba2}, making the analysis of normal vector features more challenging. Considering these factors, this paper extracts the skeleton point cloud, representing the main structure of the environment. Furthermore, it utilizes a fast range image based approach to extract normal vector features, enabling a faster and more accurate representation of the scene's degeneracy.

\section{SYSTEM OVERVIEW}

The algorithm we suggest adheres to the general framework shown in Fig. \ref{overview}. This framework comprises two primary segments: point cloud filtering and 3D LiDAR odometry. Operating on the foundation of LOAM, this system permits substituting the LO portion with various mainstream LiDAR odometry methods. The primary enhancement lies in the filtering module, where the conventional fixed-threshold clustering technique is replaced with ambient-aware filtering. The intention behind this upgrade is to furnish more accurate and consistent point cloud data for subsequent modules. The ambient-aware filtering comprises two pivotal components:

\begin{enumerate}
\item{\textit{Ambient Skeleton Extraction: }After removing the ground using a ground filter, the point cloud is subjected to range image based Euclidean and depth clustering. This process assigns dual labels to each point cloud. By analyzing the clustering scales and structural features, the algorithm extracts the ambient skeleton point cloud, enabling the reconstruction of the scene's fundamental structure.}
\item{\textit{Normal Vector Features Based Degeneration Detection: }The algorithm extracts the normal vector features of the 3D point cloud based on the range image. By analyzing the structural characteristics of the normal vector field, it evaluates the degree of degeneration in the scene. This evaluation is further mapped to dynamic weights for the depth clustering process.}
\end{enumerate}

In the actual implementation process, we embedded the proposed algorithm into the LeGO-LOAM framework. However, we believe that this module can be integrated into other LOAM-based SLAM frameworks to enhance system stability.

\begin{figure}[!t]
\centering
\includegraphics[width=3.3in]{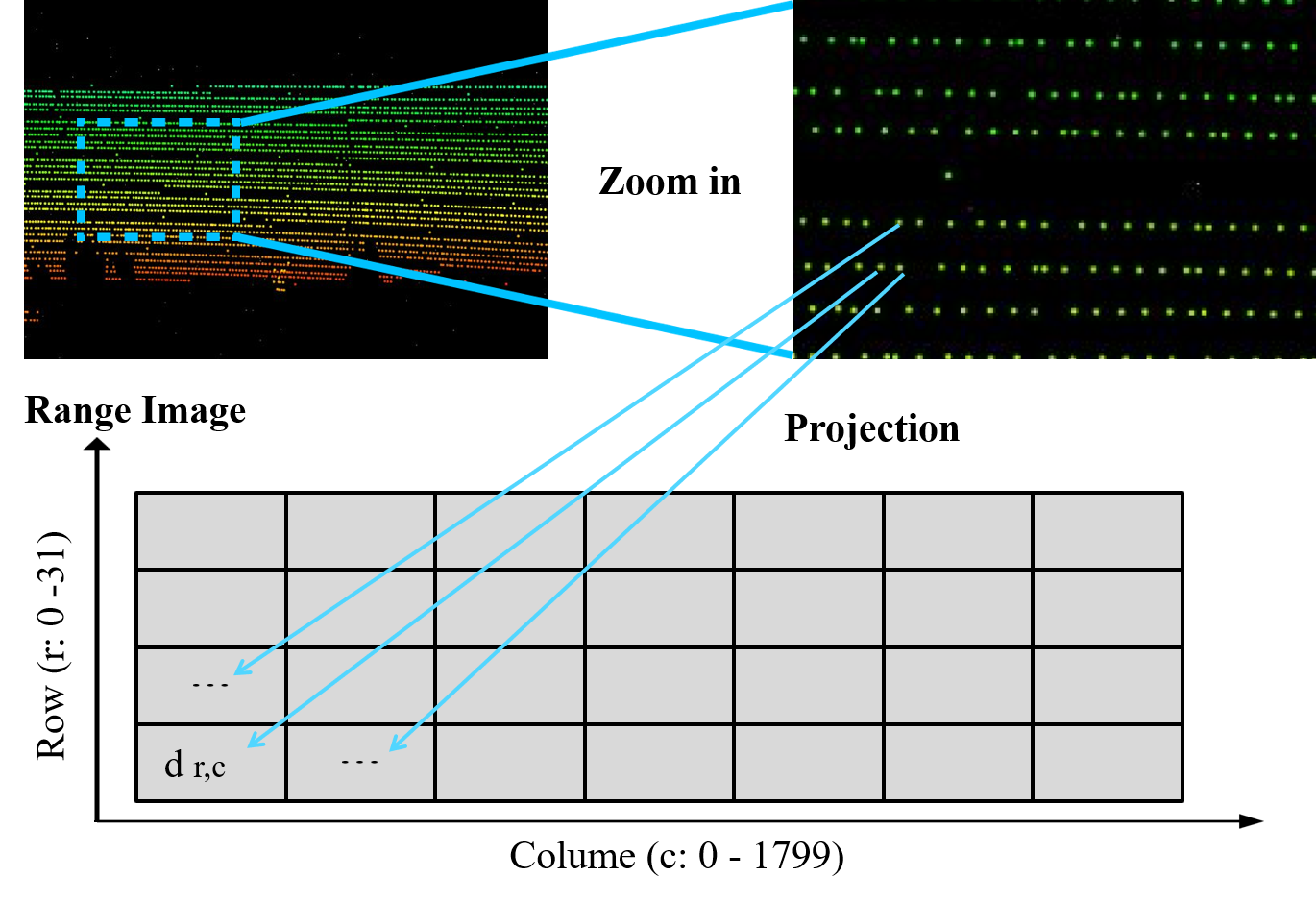}
\caption{The process of projecting the collected point cloud into a 2D range image is achieved through the physical characteristics of the 32-line LiDAR sensor.}
\label{projection}
\end{figure}

\section{AMBIENT SKELETON EXTRACTION}
3D point cloud information is noisy, and we define points that can reflect scene structure information as ambient skeleton points. Extracting ambient skeletons is mainly divided into three parts: depth clustering, range image based adaptive Euclidean clustering, and skeleton point cloud extraction.

\subsection{Depth Clustering Summary}

Depth Clustering\cite{depth} is a common CCL-based clustering method, which LeGO-LOAM utilizes in the point cloud preprocessing stage. It rapidly segments and removes small clusters, eliminating trivial and unreliable features like leaves and other small artifacts from the point cloud.

By utilizing the geometric characteristics of the point cloud collected by the LiDAR sensor, we first project it onto a 2D range image. As shown in Fig. \ref{projection}, this is a common process where the 32-line LiDAR point cloud is mapped onto a range image. The range image consists of 32 scan lines, each containing 1800 laser points. On the range image, weak adjacency exists between adjacent laser points, meaning that the nearest points are always located within a neighborhood range of a particular point. When neighboring laser points fall on the same surface, they are strictly the nearest points to each other.

\begin{figure}[!t]
\centering
\includegraphics[width=3.3in]{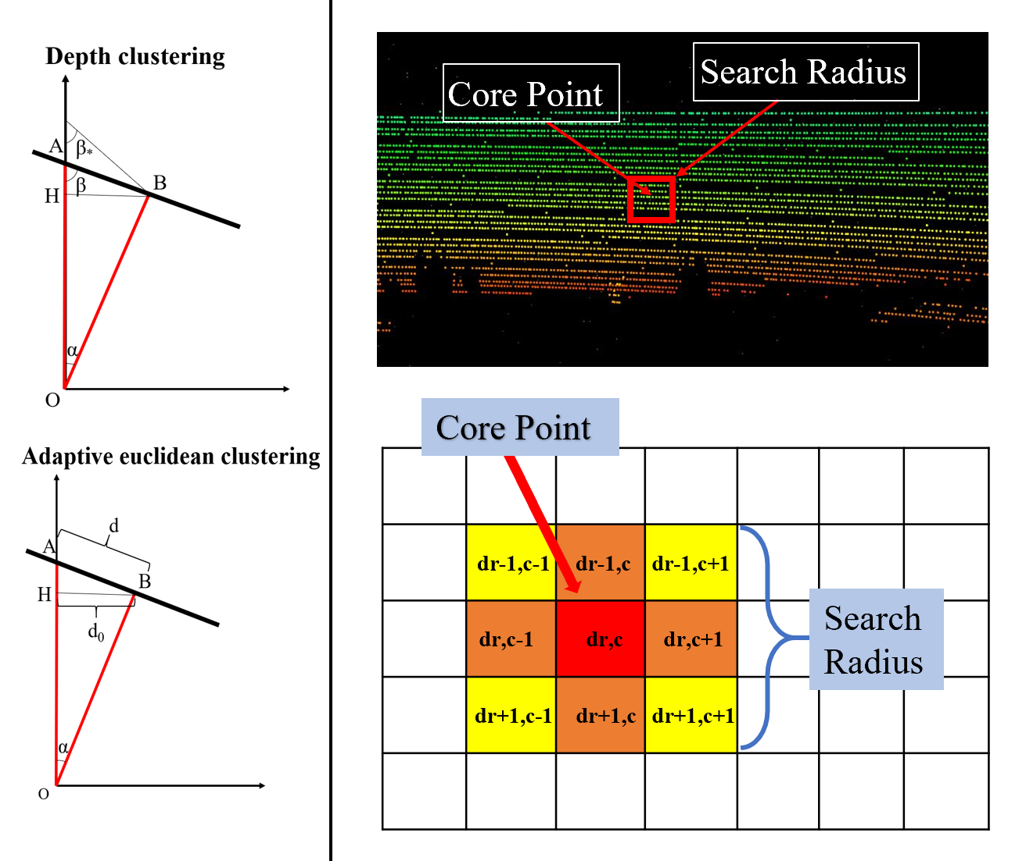}
\caption{Illustrations of the two range image based clustering methods.
 Depth clustering uses $\beta > \beta_0$ as the segmentation condition, with the search range limited to directly adjacent points, indicated by the orange area. On the other hand, Euclidean clustering uses $d < \gamma d_0$ as the segmentation condition, with the search range confined within the yellow search box.}
\label{clustering}
\end{figure}

As shown in Fig. \ref{clustering}, the orange dots represent the data points that are directly adjacent to the selected point $(r,c)$, and they are also the points in the neighborhood that are considered for each iteration of depth clustering. The core of depth clustering lies in using the angle $\beta$ between the connecting line of adjacent point clouds and the laser beam as the segmentation criteria for clustering. Considering the characteristics of the laser range measurements, we have the distance $\|OA\|$ corresponding to the first laser measurement and $\|OB\|$ corresponding to the second laser measurement. With this information, we can calculate the angle $\beta$ by applying trigonometric equations. The angle $\beta$ can be calculated using the following formula:

\begin{equation}
\label{thres1}
\beta=\arctan \frac{\|OB\| \sin \alpha}{\|OA\|-\|OB\| \cos \alpha}
\end{equation}

\begin{figure}[!t]
\centering
\subfloat[]{\includegraphics[width=1.15in]{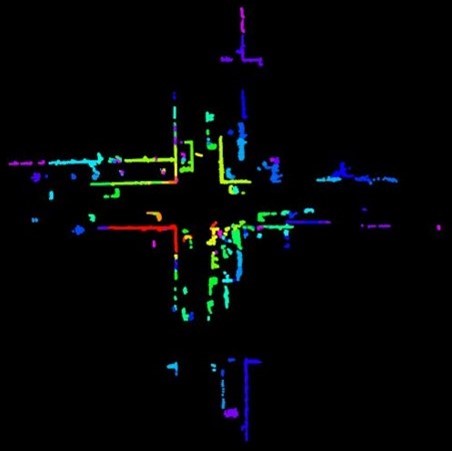}%
\label{depth_10}}
\hfil
\subfloat[]{\includegraphics[width=1.15in]{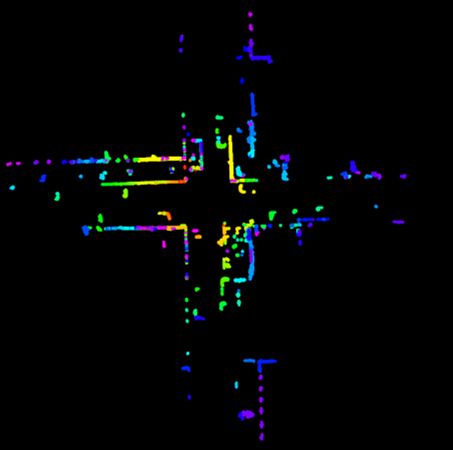}%
\label{depth_30}}
\hfil
\subfloat[]{\includegraphics[width=1.15in]{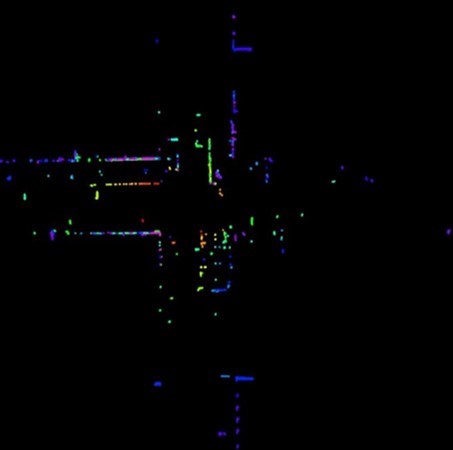}%
\label{depth_60}}
\caption{The effect of depth clustering on cleaning the point cloud at different threshold values. (a) Result with $\beta_0=10^{\circ}$: It retains the majority of the point cloud data. (b) Result with $\beta_0=30^{\circ}$: It filters out some points relative to $\beta_0=10^{\circ}$, but the overall effect is similar. (c) Result with $\beta_0=60^{\circ}$: It mainly preserves the structural information of the point cloud.}
\label{fig_sim}
\end{figure}

Depth clustering is sensitive to depth variations. When the angle $\beta$ is greater than the threshold value $\beta_0$, it indicates that the depth difference between adjacent points is small, and they are considered to belong to the same cluster. Conversely, if $\beta$ is less than $\beta_0$, the points are considered to belong to different clusters. The setting of the threshold $\beta_0$ in depth clustering is crucial: as shown in Fig. \ref{fig_sim}, different threshold values result in varying effects on depth clustering and the removal of small clusters. Under strict segmentation conditions, the method can effectively extract meaningful surface structure points but may lose a substantial amount of point cloud information. On the other hand, under lenient segmentation conditions, the filtering effect is reduced, but it preserves more point cloud information. The relationship between threshold settings and point cloud quantity is not linear. When the threshold is set to $10^{\circ}$ and  $30^{\circ}$ degrees, their difference is not significant. However, when the threshold reaches $60^{\circ}$, the difference between them becomes very noticeable.

While depth clustering displays notable prowess in point cloud cleaning, its accuracy in segmenting objects can be compromised, leading to potential misclassification of the same object into distinct clusters. Consequently, for object segmentation, this study abstained from utilizing depth clustering and instead embraced the sturdier Euclidean clustering approach. However, it's important to note that the substantial structural point extraction capabilities of depth clustering persist in their application during the skeleton extraction and point cloud cleaning stages.

\subsection{Range Image Based Adaptive Euclidean Clustering}

Euclidean clustering is one of the most commonly used algorithms in 3D point cloud analysis, especially in object segmentation tasks. The algorithm iterates through each point in the point cloud and examines neighboring points within a specified radius. If the distance between two adjacent points is below a certain threshold, they are considered part of the same cluster. This process is recursive and propagates along connected neighboring points until all points in the cluster are identified.

Although Euclidean clustering exhibits good segmentation characteristics, its application in 3D LiDAR SLAM is not extensive due to two main reasons. Firstly, Euclidean clustering is sensitive to point cloud density, and the density of point clouds collected by LiDAR varies with the depth of laser points. Using a fixed threshold cannot effectively segment objects at different depths. Secondly, traditional Euclidean clustering employs the KDTree algorithm to find neighboring points, and frequent nearest neighbor queries can lead to significant computing cost, which is unsuitable for real-time SLAM applications that demand high efficiency.

Addressing these two issues, this paper recognizes the ordered nature of the point cloud acquired by LiDAR, a characteristic well-reflected in the range image. Exploiting this characteristic can accelerate the computation speed and enhance clustering accuracy.

On one hand, we adopted a neighboring point search method similar to depth clustering. As shown in Fig.  4, we consider the nearby pixels on the range image as the adjacent points during the scanning process. Unlike depth clustering, we not only judge the points right next to each other but also perform a query for neighboring points within a 2D search window on the range image. This approach makes it easier to segment the same object effectively.

On the other hand, since we directly conduct neighborhood search using the range image, we are no longer restricted by the limitations of KDTree search radius increasing with the depth of the point cloud. Therefore, we can dynamically adjust the segmentation threshold of the Euclidean clustering directly based on the depth of the point cloud. Specifically, we calculate the threshold $d_0$ for the judgment using the following formula, where the redundancy factor $\gamma$ adjusts the looseness of the judgment and is a constant slightly greater than one(1.2 in this paper) to tolerate the inherent errors of the LiDAR:

\begin{equation}
\label{thres2}
d_0 = \gamma sin (\alpha) \|OB\|
\end{equation}

From Fig.  4, it can be observed that $d_0$ is a value that increases with the depth of the laser points. When the point cloud in this region lies on the same plane, the distance between $\|AB\|$ is relatively close. By utilizing this value, we can dynamically determine the criteria for Euclidean cluster segmentation, which better adapts to the variation in point cloud density caused by changes in depth.

\subsection{Skeleton Point Cloud Extraction}
Through depth clustering and Euclidean clustering, we assign dual labels to each point cloud, and then we need to utilize the clustering information to extract the skeleton point cloud of the scene, which represents the basic structure of the scene.

\begin{figure}[!t]
\centering
\subfloat[]{\includegraphics[width=1.15in]{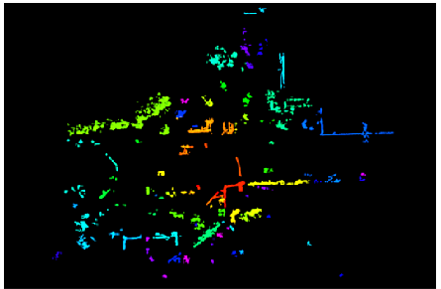}%
\label{skeleton extraction_1}}
\subfloat[]{\includegraphics[width=1.15in]{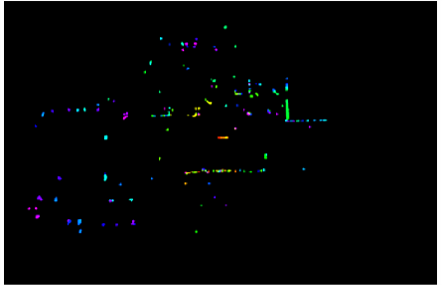}%
\label{skeleton extraction_2}}
\subfloat[]{\includegraphics[width=1.15in]{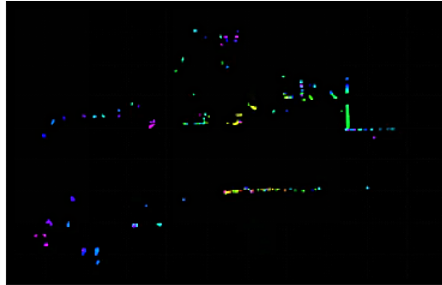}%
\label{skeleton extraction_3}}
\caption{The three stages of extracting the skeleton point cloud are as follows: (a) contains large-scale objects extracted by Euclidean clustering, (b) contains surface structural points extracted by depth clustering, and (c) is the skeleton point cloud representing the scene structure obtained by combining both clustering methods.}
\label{skeleton extraction}
\end{figure}

Firstly, we filter out small clusters for both types of clustering. Depth clustering aggregates surface structure points into larger clusters, and by using a small threshold value $n_{d}$, we can remove small clusters while preserving surface structure information. Euclidean clustering assigns the same label to points belonging to the same object, and small objects typically correspond to dynamic objects, while the point cloud of the scene is mainly composed of large objects. Therefore, by using a larger threshold value $n_{e}$, we can retain large objects. By filtering out small clusters, we obtain two sets of point clouds, each reflecting different aspects of the scene: as is shown in Fig. \ref{skeleton extraction}, depth clusters represent the scene's structure and Euclidean clusters representing large object entities. Only points with both labels are considered as valid points that truly reflect the characteristic skeleton point cloud of the scene. Specifically, the extraction method for the skeleton point cloud can be represented by Alg. \ref{alg_skeleton}. Due to the fact that the actual algorithm involves more complex data structures like maps and queues, the description provided here is a relatively concise representation of the entire extraction process.

\begin{algorithm}[!t]
\caption{Skeleton Point Cloud Extraction.}\label{normal_extraction}
\renewcommand{\algorithmicrequire}{\textbf{Input:}}
\renewcommand{\algorithmicensure}{\textbf{Output:}}
	\begin{algorithmic}[1]
        \REQUIRE $\textbf{\emph{Point Cloud}}$
        \ENSURE  $\textbf{\emph{Skeleton Point Cloud}}$
		\STATE $\{PC_{e}\}\leftarrow\ $perform Euclidean clustering on the point cloud
        \STATE $\{PC_{d}\}\leftarrow\ $perform depth clustering on the point cloud
        \FOR{PC in $\{PC_{e}\}$}
        \IF{$PC < n_{e}$}
        \STATE remove PC from $\{PC_{e}\}$
        \ENDIF
        \ENDFOR
        \FOR{PC in $\{PC_{d}\}$}
        \IF{$PC < n_{d}$}
        \STATE remove PC from $\{PC_{d}\}$
        \ENDIF
        \ENDFOR
        \STATE $PC_{E} \leftarrow aggregate\ PC\in \{PC_{e}\}$
        \STATE $PC_{D} \leftarrow aggregate\ PC\in \{PC_{d}\}$
        \FOR{point in $\textbf{\emph{Point Cloud}}$}
        \IF{$point \in PC_{E}$ and $point \in PC_{D}$}
        \STATE add point to $\textbf{\emph{Skeleton Point Cloud}}$
        \ENDIF
        \ENDFOR
        \RETURN $\textbf{\emph{Skeleton Point Cloud}}$
	\end{algorithmic}  
\label{alg_skeleton}
\end{algorithm}

At this stage, we have obtained the skeleton point cloud of the scene. These point clouds will be fed into the subsequent degeneration scene detection module, where the analysis of the skeleton point cloud will be conducted to determine the degree of scene degeneration.

\section{NORMAL VECTOR FEATURES BASED DEGENERATION SCENE DETECTION}
The degeneration detection method in this paper is primarily divided into two parts: normal vector features extraction and degeneration degree analysis.

\subsection{Normal vector features extraction}
Dense normal feature is a typical normal vector feature utilized to represent the structure of a scene. It is composed of weighted normal vectors $n_i=|n_i|\cdot\widehat{n}_i$ of the point cloud within the scene. Here, the unit normal vector $\widehat{n}_i$ reflects the geometric structure of the scene, while the weight $|n_i|$ signifies the confidence level of that vector. Extracting dense normal features enables a more concise expression of geometric information, facilitating the analysis of the degeneration degree of the scene.

Initially, due to the vast quantity of 3D point cloud data, we employ a method of randomly selecting a subset of points. This selection process aims to strike a balance between reducing computational load to achieve real-time degeneration analysis while still meeting the standards for degeneration detection. Specifically, we randomly sample a proportion $p$ (where $p<1$, and in experiments, it's set at 10\%) of points for the purpose of normal vector computation.

Next, we compute the normal vectors for the selected point cloud. Computing the normal vector of a point requires utilizing spatial information from other points within its neighborhood. This necessitates searching within a specific range of neighboring points. Using a KDTree-based method for searching becomes impractical due to its high time complexity, which fails to meet the real-time mapping demands of SLAM. Consequently, we opt to use the adjacency information of the point cloud constructed from the range image. By traversing the range image and assuming the selected point's 2D coordinates are $(r, c)$, we consider points $\{(r_n, c_n)| r_n\in[r-w, r+w], c_n\in[c-w, c+w]\}$ in the range image as neighbors, where $w$ represents the window size. We assess whether the depth difference between each neighboring point within the search window and the selected point is smaller than a threshold $depth_0$. If this condition is met, the neighboring point is added to the point cloud set $PC_n$. If the number of points in $PC_n$ is smaller than the threshold $t_0$, the point is considered an outlier and is discarded.

Subsequently, for the point cloud neighborhood set $PC_n$, we employ the Principal Component Analysis(PCA) method to calculate the directions of the normal vectors. Firstly, we perform a centering operation on the point cloud, which involves subtracting the coordinates $(x_i, y_i, z_i)$ of each point within the point cloud by the centroid coordinates of the entire point cloud. This results in a centered point cloud $PC'_n$ with its centroid at the origin. Specifically, for each point's coordinates $(x_i, y_i, z_i)$, the following calculation is performed:

\begin{equation}
\begin{aligned}
\label{gravity_core}
x^{\prime}_i=x_i-\frac{1}{N}\sum_{j=1}^{N}x_j\\
y^{\prime}_i=y_i-\frac{1}{N}\sum_{j=1}^{N}y_j\\
z^{\prime}_i=z_i-\frac{1}{N}\sum_{j=1}^{N}z_j
\end{aligned}
\end{equation}

In the second step, we compute the covariance matrix $\Sigma$ for all points in $PC'_n$ to determine the eigenvectors $\boldsymbol{v}$ and eigenvalues $\lambda$, where $(x_i, y_i) \ and \ (x'_i, y'_i)$ represent the same point in the two distinct coordinate systems of $PC_n \ and \ PC'_n$.

\begin{equation}
\label{eigenvalue1}
\begin{aligned}
\Sigma & =\frac{1}{N_{i}} \sum_{j=1}^{N_{i}} p_{j}^{\prime} p_{j}^{\prime T}
\end{aligned}
\end{equation}

\begin{equation}
\label{eigenvalue2}
\begin{aligned}
\Sigma \boldsymbol{v} & =\lambda \boldsymbol{v}
\end{aligned}
\end{equation}

In a 3D system, the covariance matrix has three eigenvalues. The plane formed by the eigenvectors corresponding to the largest and second largest eigenvalues represents the fitted plane within the point cloud set $PC_n$. The eigenvector $\boldsymbol{v}_{min}$ corresponding to the smallest eigenvalue $\lambda_{min}$ is orthogonal to this plane and serves as the required unit normal vector that we want:

\begin{equation}
\label{normal verctor1}
\begin{aligned}
\widehat{n}_i=v_{\min}
\end{aligned}
\end{equation}

Next, we calculate the magnitude of the normal vector denoted as $|n_i|$. The magnitude $|n_i|$ is determined by the number of points $N_i$ included in the neighborhood set and the measured depth $d_i$ of the scanning point. When $d_i$ is shorter and $N_i$ is larger, it indicates that points are denser and the planarity is greater, resulting in more pronounced point cloud features and a higher confidence level for the normal vector. We use the following equation to calculate the reliability of the normal vector feature:

\begin{equation}
\label{normal verctor2}
\begin{aligned}
|n_i|=\ln((d_{max}-d_i)N_i+1)
\end{aligned}
\end{equation}

By following the aforementioned steps, we have obtained the normal vectors:

\begin{equation}
\label{normal verctor field}
\begin{aligned}
n_i=|n_i|\cdot\widehat{n}_i
\end{aligned}
\end{equation}

The specific demonstration is illustrated as shown in Alg. \ref{normal_extraction}. By examining the set of normal vectors \textbf{\emph{n}} of the environment, we can reflect the geometric information of the surroundings. However, in order to quantitatively analyze the degeneration degree of the environment, we need to further analyze the dense normal features and compute specific degeneration degree values.

\begin{algorithm}[!t]
\caption{Normal Vector Features Extraction.}\label{normal_extraction}
\label{alg1}
    \renewcommand{\algorithmicrequire}{\textbf{Input:}}
	\renewcommand{\algorithmicensure}{\textbf{Output:}}
	\begin{algorithmic}[1]
        \REQUIRE $\textbf{\emph{Range Image, p}}$
        \ENSURE  $\textbf{\emph{n}}$
		\STATE R $\leftarrow$ $\textbf{\emph{Range Image}}$
		\FOR {r = 1 ... $R_{rows}$}
        \FOR{c = 1 ... $R_{cols}$}
        \STATE random accept R(r,c) with probability $p$
        \STATE $PC_n\leftarrow\{\}$
        \FOR{\{$r_n$,$c_n$\}$\in$ Neighborhood of \{r,c\}}
        \STATE depth $\leftarrow|R(r_n,c_n)|$
        \IF{$depth < depth_0$}
        \STATE add the point at \{$r_n$,$c_n$\} to $PC_n$
        \ENDIF
        \ENDFOR
        \IF{num of points in $PC_n>t_0$}
        \STATE $PC'_n\leftarrow $centering the points in $PC_n$
        \STATE get $\Sigma,\lambda,\boldsymbol{v}$ of $PC'_n$
        \STATE $\widehat{n}_i=\boldsymbol{v}_{min}$
        \STATE $|n_i|=\ln((d_{max}-d_i)N_i+1)$
        \STATE $n_i=|n_i|\cdot\widehat{n}_i$ 
        \STATE n += $n_i$
        \ENDIF
        \ENDFOR
        \ENDFOR
        \RETURN $\textbf{\emph{n}}$
		
	\end{algorithmic}  
\end{algorithm}

\begin{figure}[!t]
\centering
\includegraphics[width=3.5in]{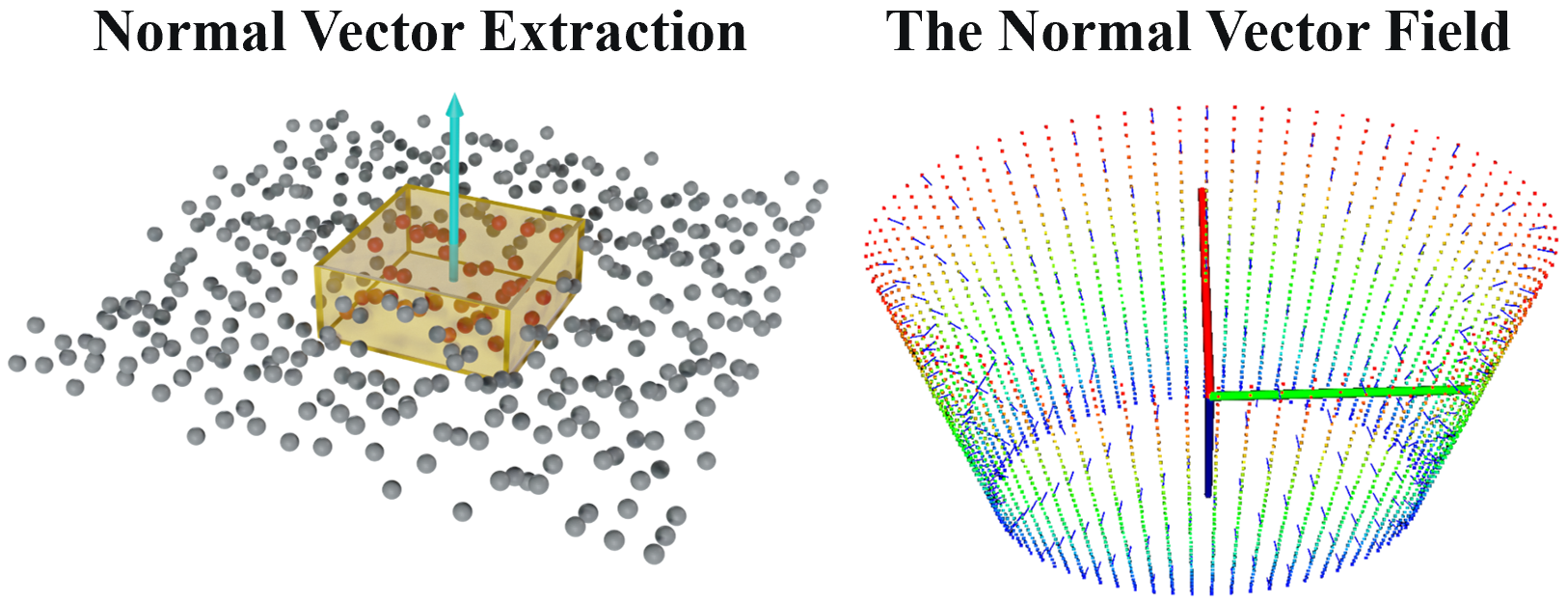}
\caption{The illustration of 3D point cloud normal vector extraction and the extracted normal vector field from the 3D scene.}
\label{normal_vector}
\end{figure}

\subsection{Degeneration Degree Analysis}

In the previous step, we obtained the normal vector features. Due to subsequent modules optimizing the ground points along the z-axis within the framework of this paper, we simplify calculations by discarding the z-coordinate of all normal vector features. As a result, the normal vector $n_i=(x_i, y_i, z_i)$ transitions from three-dimensional to two-dimensional form, represented as $v_i=(x_i, y_i)$.

Moving forward, we utilize the obtained new normal vectors $v_i$ to calculate the degeneration degree. In order to reduce the dimensionality of the high-order features, we employ the PCA method to obtain eigenvectors and eigenvalues. Among these, the direction of the eigenvector $v_{max}$ associated with the larger eigenvalue can represent the degeneration direction of the scene.

For further calculation of geometric degeneration degree, we establish a new coordinate system, with $v_{max}$ as the x-axis and the eigenvector $v_{min}$ associated with the smaller eigenvalue as the y-axis. We transform all the normal vectors $v_i$ into the new coordinate system, denoted as $v_i'$. We project all vectors in $v'$ onto the coordinate axes, obtaining their components along the axes. By calculating the ratio $k$ between the x-axis component and the y-axis component, we quantitatively reflect the degree of scene degeneration:

\begin{equation}
\label{get_k}
\begin{aligned}
k=\frac{\sum_{i=1}^{M}{x^{\prime}}_i}{\sum_{i=1}^{M}{y^{\prime}}_i}
\end{aligned}
\end{equation}

The range of $k$ is $[1, \infty]$, and we map it to the interval $[0, 1]$ to obtain the degeneration degree $\mu$. A value of $\mu$ closer to $1$ signifies a higher degree of degeneration. The mapping function we employ is as follow: 

\begin{equation}
\label{get_mu}
\begin{aligned}
 \mu = -\frac{1}{k} + 1 \
\end{aligned}
\end{equation}

In this mapping function, when $\mu$ is just above 1, the degeneration degree increases rapidly, while as it approaches $\infty$, the growth of the degeneration degree becomes slower. When the ratio of different components reaches a certain threshold, the degeneration degree essentially stops changing, which aligns with practical scenarios.

When performing depth clustering to clean the point cloud, the decision of whether two points belong to the same cluster depends on a dynamic threshold value $\beta_0$, with upper and lower bounds $\beta_{0}^{max}$ and $\beta_{0}^{min}$ respectively. As the degeneration degree $\mu$ increases, indicating greater environmental degeneration, the threshold value chosen during clustering should be minimized, $\beta_{0}^{min}$ is selected. Conversely, when $\mu$ decreases, implying lower environmental degeneration, $\beta_{0}^{max}$ is chosen. Therefore, we map the range of the degeneration degree $\mu$ from $[0, 1]$ to the clustering threshold range using the following mapping function:

\begin{equation}
\label{map_beta}
\begin{aligned}
\beta_0 = \mu(\beta_{0}^{min} - \beta_{0}^{max}) + \beta_{0}^{max}
\end{aligned}
\end{equation}

In the end, we will use the obtained dynamic segmentation threshold to perform another round of depth clustering on the point cloud, ultimately obtaining a point cloud that retains as much information as possible while being sufficiently clean.

\section{EXPERIMENTS}
\subsection{Experiment Setup}
In this section, the proposed algorithm was experimentally validated both on a dataset and in a real environment. The hardware platform used for testing was the AMD R7 5800H CPU, and the software platform used was the robot operating system (ROS) version noetic. 

The testing was firstly performed on the KITTI benchmark, which is one of the most commonly used dataset in the field of autonomous driving. The KITTI benchmark captures data using the HDL-64E LiDAR at a frequency of 10 Hz and includes a large amount of real driving scene data. The experiments in this section focused on testing point cloud segmentation, degeneration scene detection, and overall localization accuracy, with detailed analyses provided for each test. Furthermore, this article also conducted practical experiments to evaluate the mapping performance. The experimentation utilized a VLP-16 LiDAR operating at a 10 Hz frequency. The primary objective of these real-world tests was to demonstrate the system's robustness and underscore the essential nature of the point cloud filtering mod ule.

\subsection{Segmentation Performance}

We first compared the effects of different clustering on the KITTI benchmark. Instead of using the original annotations from the KITTI dataset, we utilized the re-annotated KITTI dataset provided by Yang et al\cite{seg-bench}. For the KITTI dataset, they chose 200 frames from 3D object detection data and re-annotated them. This was because the original bounding boxes are based on RGB image projection, which includes estimated full vehicle size. As a result, these boxes don't objectively showcase the clustering method's performance. In Table \ref{tab1}, a comprehensive comparison is made between our algorithm and several widely-used 3D point cloud clustering algorithms in terms of accuracy and time efficiency. Among the algorithms evaluated, two were based on raw point clouds: the popular Euclidean clustering and Density-Based Spatial Clustering(DBSCAN). Moving to CCL-based clustering, we analyzed a number of representative algorithms in this category, including depth clustering\cite{depth}, Run clustering\cite{run}, CCL-based Euclidean clustering, and our innovative adaptive CCL-based Euclidean clustering. It's noteworthy that only Run clustering has an inherent ground removal feature. For all other algorithms, a ray ground filter\cite{rgf} was used for this purpose. We ensured optimal performance by meticulously selecting the parameters for each algorithm through rigorous experimentation. To gauge the segmentation prowess of each method, we calculated the 3D Intersection over Union (IoU) comparing the resultant clusters to the ground-truth boxes.

Regarding accuracy, Euclidean clustering significantly outperformed other algorithms. In contrast, DBSCAN demonstrated subpar results. The observed underperformance is likely due to the inherent characteristics of scanning lasers, which exhibit varying resolutions in the horizontal and vertical axes. Consequently, this leads to a non-uniform density distribution across a single object. While Euclidean clustering based directly on raw point clouds showcases commendable accuracy, its per-frame execution speed surpasses 400ms. Given that LiDAR typically operates at frequencies up to 10Hz, a considerable amount of data is discarded, which has profound implications on the accuracy of LiDAR odometry. In fact, any clustering method based on raw point clouds inevitably involves a neighborhood query step, so other algorithms utilizing raw point clouds also face the issue of significant time consumption.

Within the realm of CCL-based algorithms, the time complexity for all is in the millisecond range, making them apt for real-time applications. Among these, the Euclidean clustering within the CCL-based framework initially displayed preeminent object segmentation results. Even in situations that demanded weak neighbor relation searches on range images, this method surpassed other popular CCL-based approaches, such as depth clustering and run clustering. Notably, our proposed adaptive threshold-enhanced algorithm surpassed the fixed-threshold Euclidean clustering, registering as the only algorithm with an IoU exceeding 50\%. This increased object segmentation precision directly boosts algorithmic accuracy, paving the way for enhanced scene comprehension and more dependable object detection.

\begin{table*}[]
\caption{PARAMETER SETTINGS AND SEGMENTATION PRECISION OF DIFFERENT METHODS ON KITTI DATASET}
\label{tab1}
\renewcommand{\arraystretch}{1.5}
\setlength{\tabcolsep}{7pt}
\centering
\begin{threeparttable}
\begin{tabular}{c|ccc|cc}
\hline
Approach                       & Type                   & Ground removal     & Clustering theta                         & Precision                        & Runtimes \\ \hline
Euclidean clustering           & Raw point clouds based & Ray ground filter & $Eps=0.75m$ & 67.1\% & 450.5ms \\
DBSCAN clustering              & Raw point clouds based & Ray ground filter & $Eps=0.75m$, $Min_{samples}=5$ & 22.5\% & 405.5ms \\
\cline{1-6} 
Run clustering                 & Range image based      & $Params_{GPF}$    & $Params_{SLR}$ & 29.3\% & 32.5ms \\
Depth clustering               & Range image based      & Ray ground filter & $\beta=10^\circ$ & 41.1\% & 1.43ms \\
CCL-based Euclidean clustering & Range image based      & Ray ground filter & $Eps=0.75m$ & 43.1\% & 4.12ms \\
\textbf{Ours}                  & Range image based      & Ray ground filter & Adaptive & 51.5\% & 4.05ms \\
\hline
\end{tabular}
\begin{tablenotes}
    \footnotesize
    \item[*] $Params_{GPF} = \{N_{segs} = 3, N_{iter} = 3, N_{LPR} = 20, Th_{seeds} = 0.4m, Th_{dist} = 0.2m\}$
    \item[**] $Params_{SLR} = \{Th_{run} = 0.5m, Th_{merge} = 1m\}$
\end{tablenotes}
\end{threeparttable}
\end{table*}

\subsection{Degeneration Scene Detection Results}

\begin{figure*}[!t]
\centering
\includegraphics[width=5.5in]{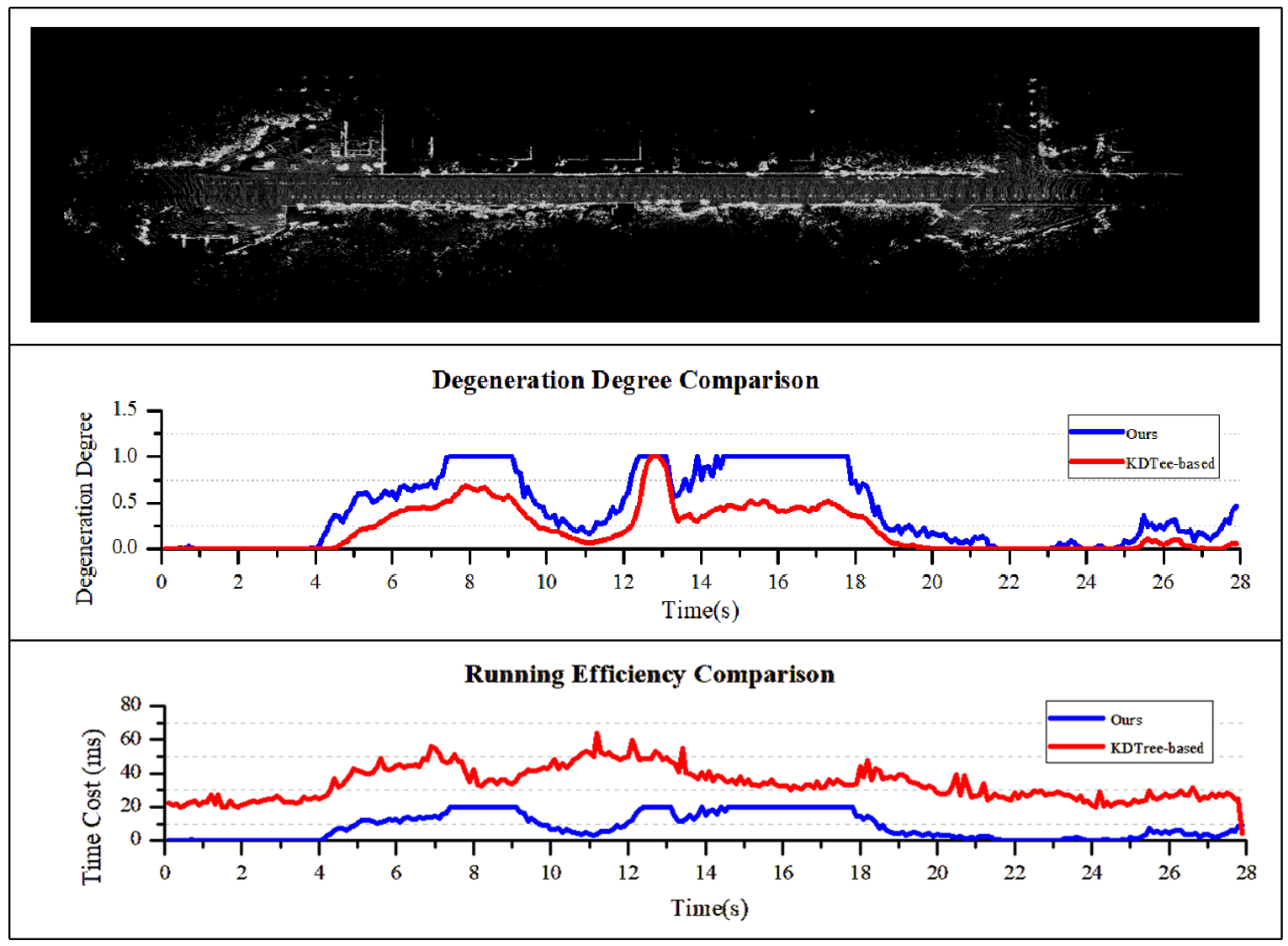}
\caption{The performance of the degeneration scene detection module was tested on the sequence 04 dataset. Red represents a generic KDTree-based degeneration scene detection method, which employs KDTree for nearest neighbor point search and performs feature extraction on all point clouds. Blue represents our algorithm, which utilizes range images for nearest neighbor point search and conducts feature extraction only on the skeleton point cloud.}
\label{degeneration_degree}
\end{figure*}

Calculating the extent of scene degeneration lacks a straightforward ground truth. In this study, we selected the sequence 04 from the KITTI dataset to visualize the detection results. Such degeneration segments are abundant in real-world scenarios and datasets. The specific segment chosen in this paper is mainly because, when combined with the constructed map, it can visually demonstrate the effectiveness of degeneration detection more intuitively. As shown in Fig. \ref{degeneration_degree},  the degeneration degrees are obviously under a low level due to the abundant features at both ends of the path. The middle parts of the long straight path, on the other hand, should experience various degrees of degeneration because of the open and structural characteristics on both sides. 

The pair of line charts presented in relation to Fig. \ref{degeneration_degree} compare our algorithm with a KDTree-based scene degeneration detection algorithm. While the conventional algorithm calculates all points within the scene, our approach utilizes range images and exclusively computes the normal vector attributes of the skeleton structure. In the line charts, the blue line represents our algorithm, and the red line represents the KDTree-based method.

As is shown in the chart, our proposed degeneration scene detection method achieves higher degeneration degrees in the degeneration sections of the road and responds faster when degeneration occurs, effectively mitigating the impact of degeneration on the environment. Although the algorithm's sensitivity improvement may not be visually apparent in terms of degeneration coefficients, in practice, the improved algorithm detects a difference of 0.5 and 1 in degeneration coefficients compared to the original algorithm in the middle range. The mapped thresholds are approximately $\beta_0=10^\circ$ and $\beta_0=40^\circ$. The algorithm presented in this paper is more likely to consider this segment as having degeneration suspicions and retains more point cloud data, thereby better avoiding degeneration occurrences. Meanwhile, the comparison algorithm takes over 60ms per frame for computation, whereas the proposed algorithm in this study completes each frame in less than 10ms. Considering that LiDAR data updates a frame every 0.1s, without optimization, computation time nearly occupies half of the processing time, which is unacceptable in practical applications. Therefore, our optimization plays a crucial role in ensuring real-time performance. Moreover, as LiDAR frequency and point cloud quantity increase, this difference will become even more pronounced.

\begin{figure}[!t]
    \centering
    \includegraphics[width=1.0\linewidth]{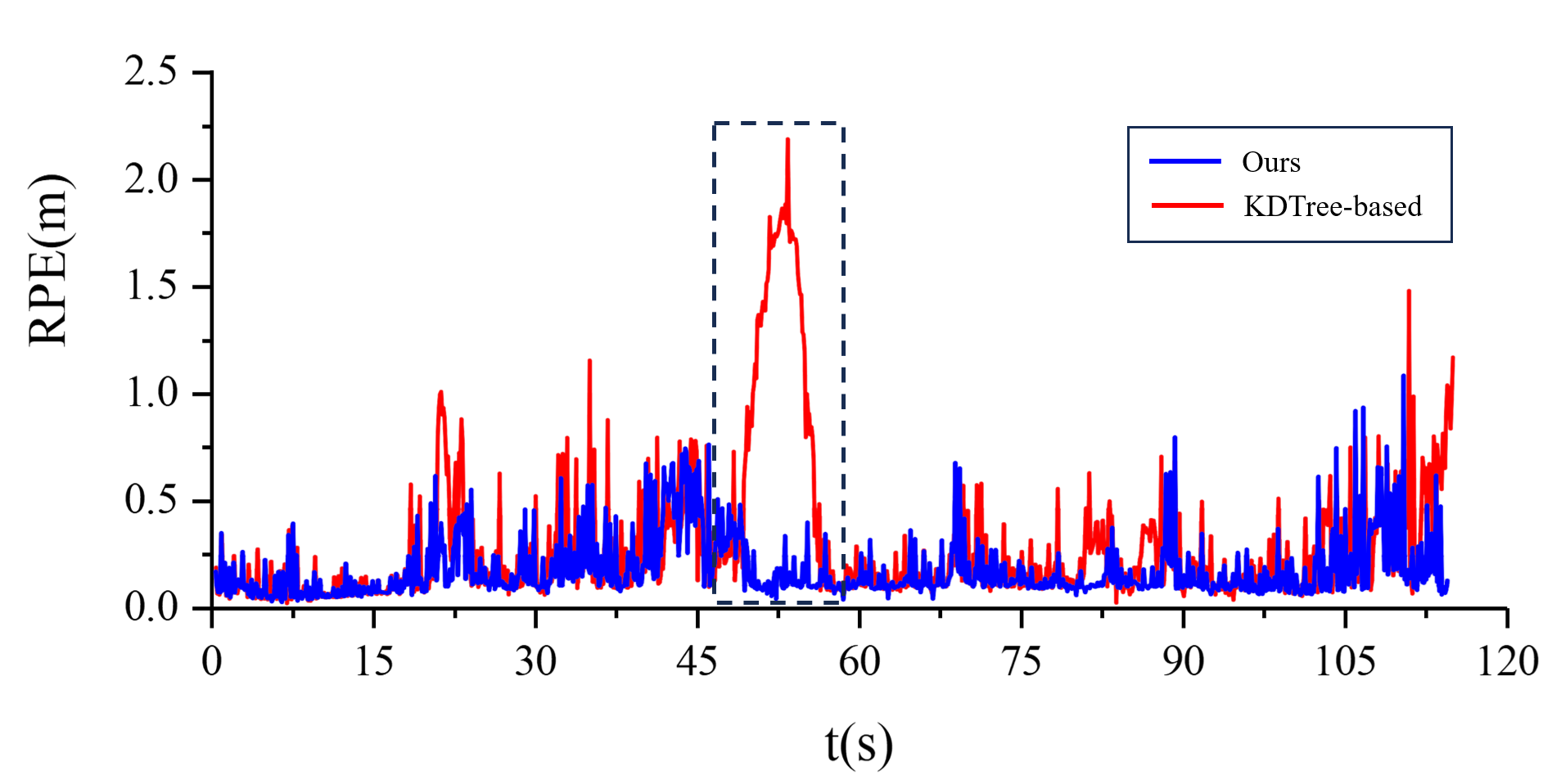}
    \caption{Comparison of the impact on localization error for sequence 01 between ours and the KDTree-based algorithm. Blue represents the relative Pose Error (RPE) over Time for Sequence 01 with the improved degeneration detection Module. Red represents the RPE using the unenhanced algorithm. The sections enclosed in black box indicates segments where degeneration occurs.}
    \label{degeneration_comp}
\end{figure}

To bolster the credibility of the enhancement, a direct comparison was conducted between the odometric localization accuracy of our advanced degeneration detection module and its predecessor. As depicted in Fig. \ref{degeneration_comp}, which represents sequence 01 of the KITTI dataset, our refined algorithm adeptly identified instances of discontinuous degeneration. Conversely, the KDTree-based version, due to its inaccurate detection results, failed to handle degeneration and led to a noticeable decrease in localization accuracy.

\subsection{Positioning Accuracy on Public Benchmark}

\begin{figure}[!t]
\centering
\includegraphics[width=3.5in]{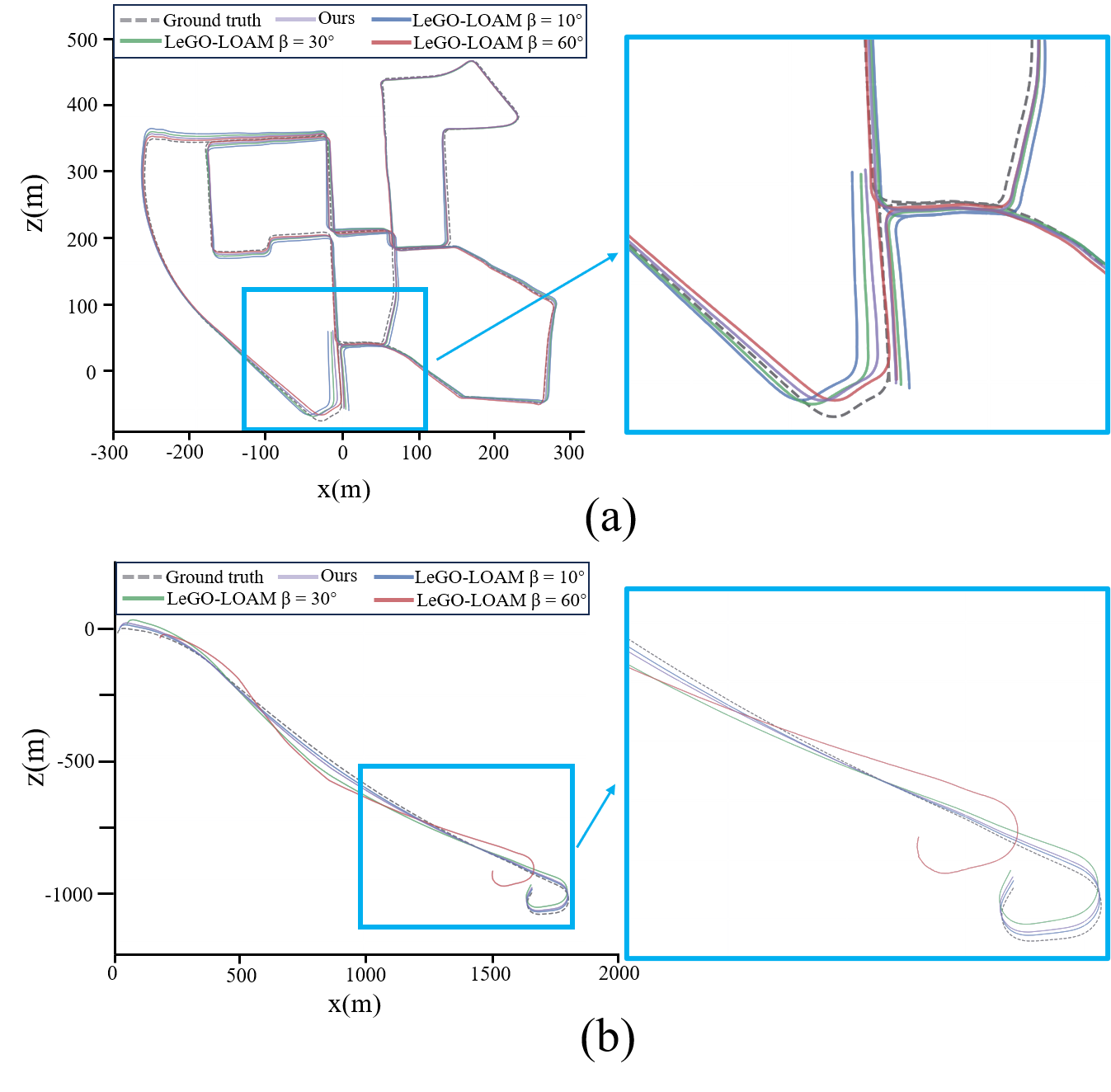}
\caption{Comparison of trajectories generated by our algorithm and LeGO-LOAM algorithm at different thresholds in sequence 00 and 01. Dashed lines represent the ground truth, purple represents our algorithm, and blue, green, and red represent trajectories with thresholds set at $\beta=10^\circ$, $\beta=30^\circ$, and $\beta=60^\circ$, respectively. (a) represents the overall trajectory of sequence 00 and a close-up of the details. (b) represents the overall trajectory of sequence 01 and a close-up of the details.}
\label{00and01}
\end{figure}

We conducted precision experiments on ten publicly available sequences from the KITTI benchmark, specifically sequences 00 through 10, excluding sequence 03 which has been removed by the officials. Our initial examination centered on sequences 00 and 01, detailed in Fig. \ref{00and01}, where our algorithm's localization accuracy is evident from the trajectory maps. The results from these sequences suggest that our algorithm competes effectively with the LeGO-LOAM algorithm, demonstrating comparable or superior performance to its optimally tuned parameters.

Then, ensuring a comprehensive understanding of the performance metrics across all sequences, we quantitatively analyzes the positioning accuracy using the Relative Pose Error (RPE) as a metric to assess the SLAM algorithm's precision. The RPE primarily describes the accuracy of the pose difference between two frames at a fixed time interval (compared to the ground truth pose), effectively measuring the LiDAR odometry error directly. As the LiDAR data's frequency is essentially constant, the time interval between each pair of frames can be considered the same. Let the algorithm's estimated poses be denoted as $\{P_i|P_1,P_2...P_n \in SE(3)\}$ and the ground truth poses be denoted as $\{Q_i|Q_1,Q_2...Q_n \in SE(3)\}$. For a time interval $\Delta t$, the RPE calculation for the i-th frame relative to the (i-1)-th frame is as follows:

\begin{equation}
\label{rpe}
\mathbf{E}_{i}=\left(\mathbf{Q}_{i}^{-1} \mathbf{Q}_{i+\Delta t}\right)^{-1}\left(\mathbf{P}_{i}^{-1} \mathbf{P}_{i+\Delta t}\right)
\end{equation}

We use the Root Mean Square Error (RMSE) to calculate the RPE and obtain an overall error value for localization, which directly measures the accuracy of a single mapping process. The calculation of the 3D SLAM's RMSE is as follows:

\begin{equation}
\label{rmse}
\operatorname{RMSE}=\sqrt{\frac{1}{n} \sum_{i=1}^{n}{E_i}^{2}}
\end{equation}

As this paper is integrated into the LeGO-LOAM framework with the main contribution being the dynamic adjustment of the point cloud segmentation and cleaning thresholds, the main reference for comparison is the performance of the LeGO-LOAM algorithm under different threshold settings. The segmentation thresholds are set at 10°, 30°, and 60°, with increasing strictness as the angle increases, resulting in more points being removed from the point cloud. Furthermore, as one of the most representative algorithms in the 3D LiDAR field, the LOAM algorithm is also used for comparison. The positioning error of LOAM exhibits an order of magnitude difference compared to other algorithms. This is because LOAM does not segment point clouds during processing but directly matches them, resulting in very slow processing times, to the extent that almost half of the laser frames cannot be processed promptly. When conducting accuracy comparisons, we employ pose interpolation to handle the missing laser frames, ensuring consistent scaling for the comparison. This illustrates the enhancement in LiDAR odometry performance with the pre-segmentation of point clouds.

\begin{table*}[!t]
\caption{DATASET VERIFICATION—ROOT MEAN SQUARE ERROR PER SCAN FOR COMPARED METHODS}
\label{tab2}
\renewcommand{\arraystretch}{1.2}
\setlength{\tabcolsep}{19pt}
\centering
\begin{tabular}{c|ccccc}
\hline
\diagbox{Sequences}{RMSEs}{Methods}        & \begin{tabular}[c]{@{}c@{}}\\ ($\beta_0=10^\circ$)\end{tabular} & \begin{tabular}[c]{@{}c@{}}LeGO-LOAM\\ ($\beta_0=30^\circ$)\end{tabular} & \begin{tabular}[c]{@{}c@{}}\\ ($\beta_0=60^\circ$)\end{tabular} & LOAM      & \textbf{Ours}               \\ \hline
Sequence 00       & 0.0816                                                    & 0.0816                                                    & \textbf{0.0738}                                           & 0.3876  & \underline{0.0751}           \\
Sequence 01       & \textbf{0.1844}                                           & 0.2238                                                    & 0.7624                                                    & 0.8872  & \underline{0.2172}           \\
Sequence 02       & 0.0806                                                    & \textbf{0.0711}                                           & 0.0788                                                     & 0.4011  & \underline{0.0779}           \\
Sequence 04       & 0.1506                                                    & 0.1355                                                    & \underline{0.0970}                                                     & 0.3875  & \textbf{0.0929}  \\
Sequence 05       & \underline{0.0600}                                                    & 0.0613                                                    & 0.0619                                                      & 0.3657  & \textbf{0.0595}  \\
Sequence 06       & \underline{0.0850}                                                    & \textbf{0.0849}                                           & 0.0883                                                    & 0.4553  & 0.0883            \\
Sequence 07       & \underline{0.0655}                                                    & \underline{0.0655}                                                    & 0.0706                                                     & 0.2765  & \textbf{0.0645}  \\
Sequence 08       & 0.0721                                                    & 0.0724                                                    & \underline{0.0720}                                                    & 0.3083  & \textbf{0.0717}  \\
Sequence 09       & \textbf{0.0669}                                            & \underline{0.0672}                                                    & 0.0857                                                    & 0.3987  & 0.0795           \\
Sequence 10      & \underline{0.0642}                                                    & \textbf{0.0617}                                           & 0.0788                                                    & 0.3124   & 0.0710           \\
Average & \underline{0.0911}                                                   & 0.0925                                                   & 0.1469                                                  & 0.3583 & \textbf{0.0898} \\ \hline
\end{tabular}
\end{table*}

As shown in Table \ref{tab2}, we computed the RMSE for different algorithms on the ten sequences of the KITTI dataset and took their average values. Bolded fonts in the table indicate the best results, while underlined fonts indicate the second-best results. From the table, it is evident that the LeGO-LOAM based algorithms demonstrated excellent localization accuracy in most sequences. Our algorithm consistently outperformed or was close to the results obtained using the optimal threshold settings. Meanwhile, our algorithm achieved the highest average accuracy. The main contribution was due to significant differences observed in sequence 01 and sequence 04: in KITTI sequence 01, LeGO-LOAM only performed well with a $\beta_0=10^\circ$ threshold, but produced severe errors with other thresholds; while in sequence 04, the LeGO-LOAM algorithm's accuracy improved as the threshold increased. In contrast, our algorithm dynamically adjusted the threshold based on scene characteristics, ensuring good results in both of these scenarios.

\begin{figure}[!t]
\centering

\includegraphics[width=3.5in]{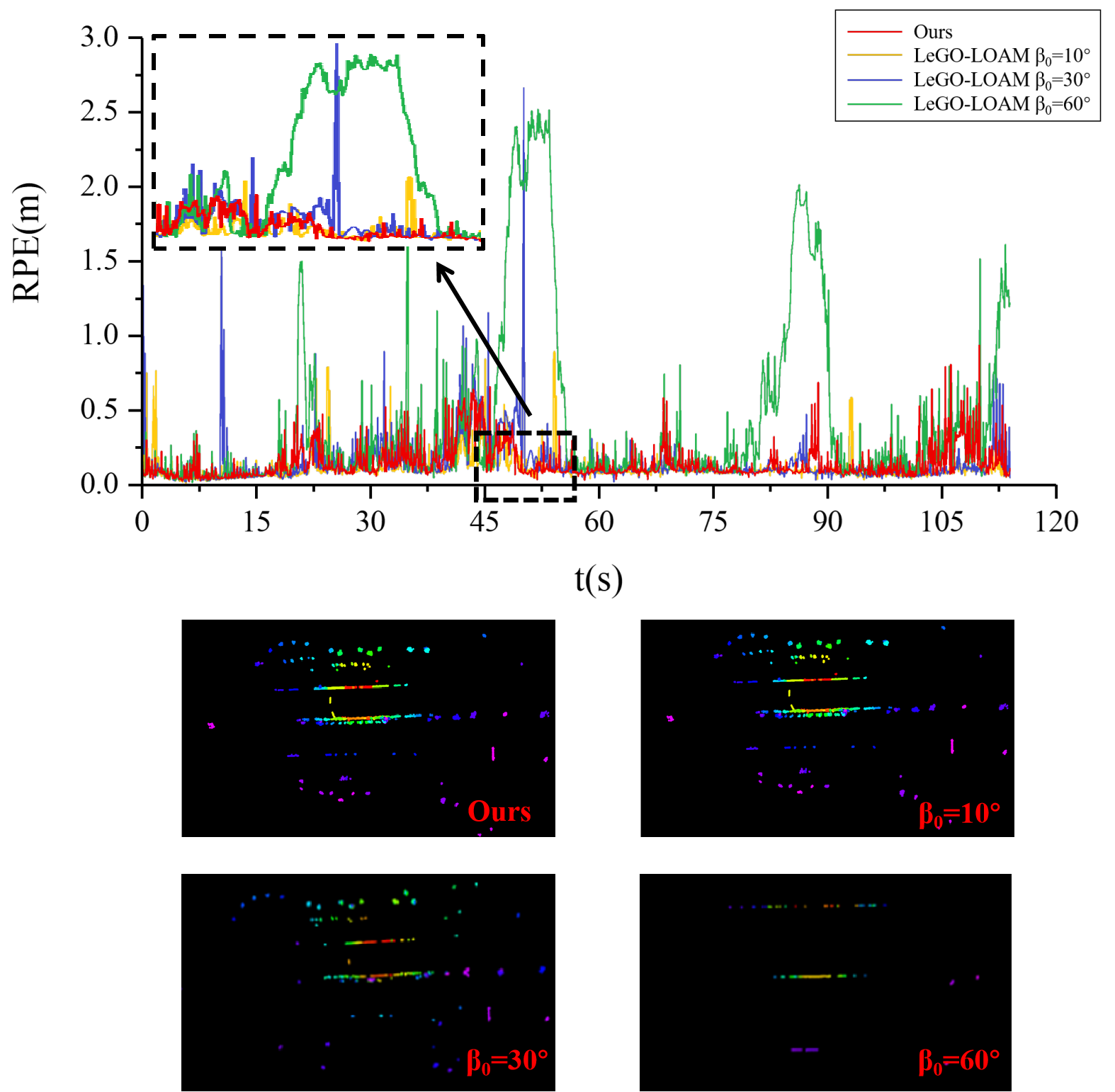}

\caption{The chart above presents the real-time localization accuracy of the algorithm at each moment along with the point cloud utilized for matching in sequence 01. The images below showcase the collected point cloud information retained within the annotated black-box areas, under varying threshold settings.}
\label{combine_01}
\end{figure}

Let's first analyze the situation in sequence 01, where LeGO-LOAM produced an RMSE of 0.76 with a $\beta_0=60^\circ$ threshold, indicating severe localization errors. As shown in Fig. \ref{combine_01}, significant errors occurred in the middle part of the trajectory, which is highlighted in a black box. It is clear that LeGO-LOAM exhibited serious localization errors when using strict threshold settings. To further analyze the issue, we need to search for answers within the scene. We extracted the point cloud information used for matching in the frames with large matching errors, as shown in Fig. \ref{combine_01}. In sequence 01, the main scene was the vehicle driving on a highway, with vegetation only present on both sides of the road. From the error plots at each moment, it is evident that when the threshold was set too large, feature point extraction became overly strict, only extracting information from the fences on both ends while discarding the point cloud data from the outer shrubs, resulting in severe degeneration issues. As a result, the robot continuously believed that it was not moving forward and couldn't accurately localize and navigate. Conversely, when the threshold was set smaller, the filtering module introduced vegetation point cloud data, thereby ensuring consistent matching.

\begin{figure}[!t]
\centering

\includegraphics[width=3.5in]{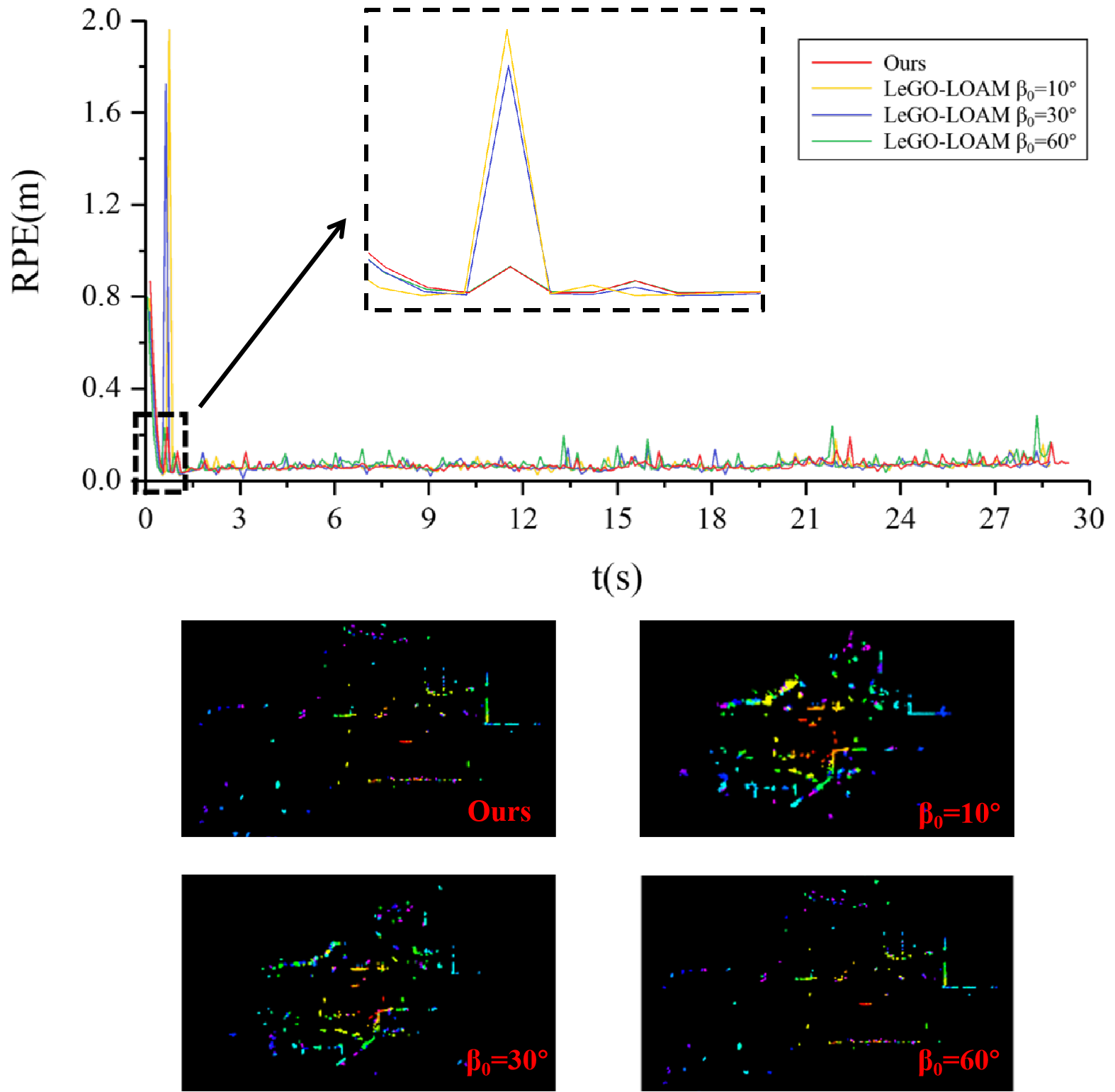}

\caption{The chart above presents the real-time localization accuracy of the algorithm at each moment along with the point cloud utilized for matching in sequence 04. The images below showcase the collected point cloud information retained within the annotated black-box areas, under varying threshold settings.}
\label{combine_04}
\end{figure}

Next, let's analyze sequence 04, where LeGO-LOAM also exhibited significant differences under different thresholds. By observing the RPE at each moment in Fig. \ref{combine_04}, we found that the error difference did not exist throughout the entire trajectory but originated from the errors at the beginning of map construction. As shown in Fig. \ref{combine_04}, we also extracted point cloud information obtained by different algorithms in the same scene which is highlighted in a black box. In sequence 04, the scene was a complex urban environment, characterized by a chaotic and cluttered environment with numerous pedestrians, vehicles, shrubs, and various small obstacles. Due to significant point cloud noise, when the threshold was set too small, the LeGO-LOAM algorithm extracted many unreliable feature points, further increasing matching errors. On the other hand, when the threshold was set larger, the algorithm enhanced its point cloud filtering ability, utilizing the essential structural points of the scene for matching, and improved the matching accuracy.

Although our analysis reveals a slight improvement in algorithm accuracy for algorithms in this article, observing the data shows that under strict threshold settings ($\beta_0=60^\circ$), the performance of most sequences is inferior to the lenient threshold. However, this does not imply that employing strict threshold settings is entirely insignificant. Scenes like that of sequence 04 are actually quite common in practical applications. Given that a significant portion of the KITTI dataset comprises relatively uncomplicated urban environments with smooth driving conditions, lacking turbulence, the point cloud data is comparatively clean. This situation presents an overly idealized scenario for real-world applications.In the next phase, we conducted experiments in real-world scenarios to reflect the necessity of point cloud data cleaning in actual environments.

\subsection{Real-world Mapping Results}

\begin{figure*}[!t]
\centering

\includegraphics[width=7in]{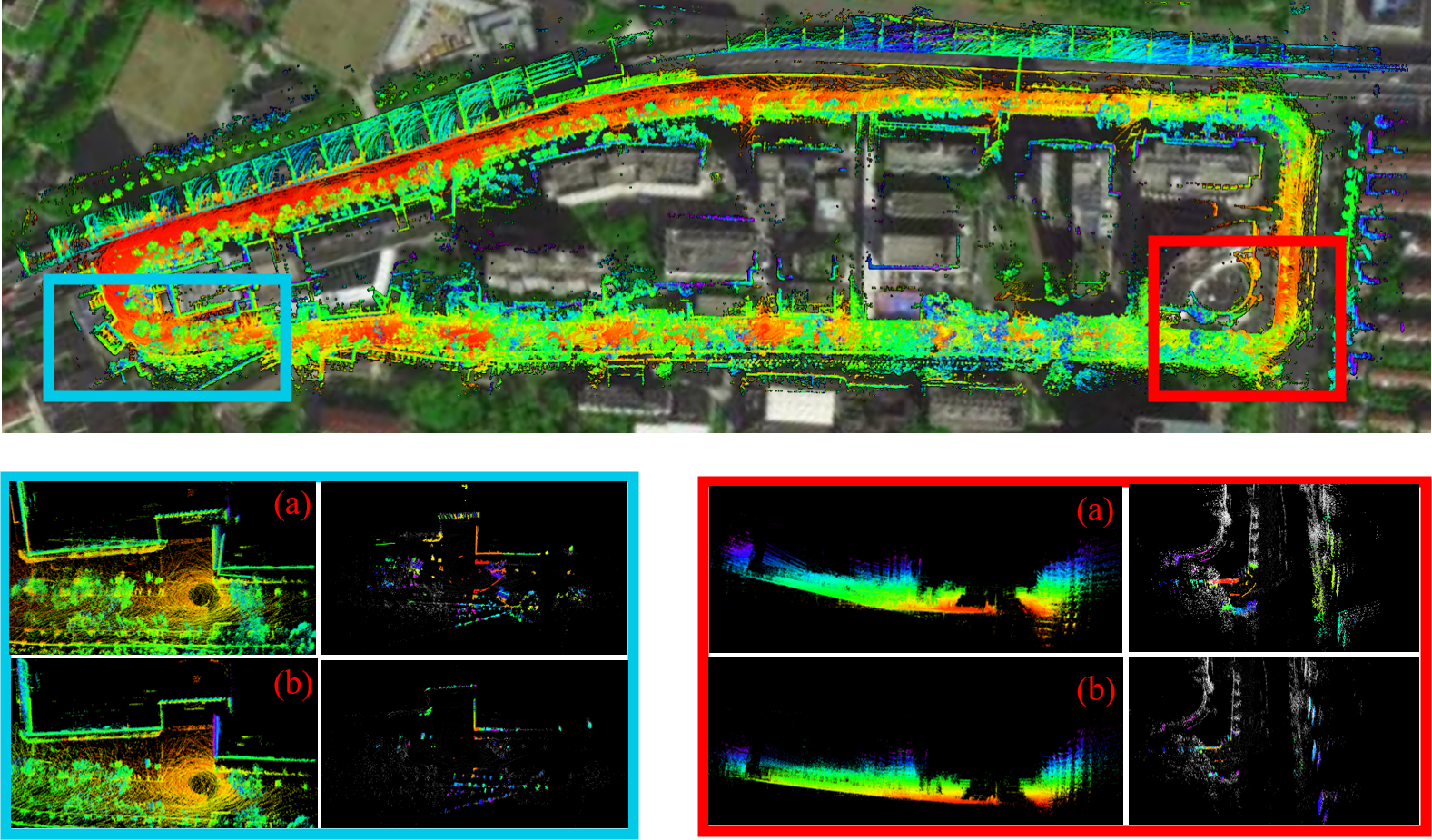}

\caption{Real-world SLAM mapping results presentation and comparative analysis of local mapping under different point cloud cleaning levels. (a) represents the LeGO-LOAM algorithm using a lenient threshold ($\beta_0=10^{\circ}$ or $\beta_0=30^{\circ}$), which retains some unreliable point cloud information, resulting in mapping errors. (b) employs the algorithm proposed in this paper, which calculates a strict threshold based on the surrounding environment. This effectively removes noise points from the point cloud, ensuring algorithm stability.}
\label{real-world}
\end{figure*}

In this section, we conducted real-world experiments within an urban environment. The chosen route formed a loop, encompassing various influential factors such as buildings, shrubs, trees, pedestrians, vehicles, and more. The data collection accounted for common factors found in real driving scenarios, including stops, starts, and LiDAR turbulence. As shown in Fig. \ref{real-world}, the map constructed by the algorithm in this article does not utilize loop closures. It can be observed that when the vehicle returns to its initial position, there is minimal accumulated error. Given the complexity of the environment, the algorithm employed strict point cloud cleaning thresholds across all road segments, ensuring the stable operation of the algorithm.

We also tested using a more lenient point cloud cleaning approach ($\beta_0=10^{\circ}$ and $\beta_0=30^{\circ}$), but neither of them ensured the complete construction of the map for this route. The primary map construction errors occurred at the two locations highlighted in Fig. \ref{real-world}. Within the blue-bordered section, the road segment was turbulent, and the pre-processing module for ground removal did not effectively filter out the ground. The laser point cloud exhibited significant jitter and contained a considerable amount of noise, leading to substantial odometry drift. This effect is visible on the map, where severe ghosting is observed, and the walls are not fully aligned. In the red area, the issue arises from an abundance of unreliable point cloud information originating from bushes and trees compared to the dependable point cloud data from buildings. The lenient threshold fails to effectively eliminate these unreliable feature points. Consequently, significant drift occurs, leading to a considerable angle of unnatural rotation in the overall map due to the prevalence of unreliable information from the vegetation cover.

Through practical experiments, it becomes evident that effective point cloud cleaning plays a crucial role in real-world localization and mapping tasks. Attempting to utilize all available point cloud data can introduce numerous erroneous matches, thereby decreasing the overall system's robustness. Therefore, it is essential to selectively utilize valid information from the environment, adapting to the scene appropriately, as an indispensable process. In conclusion, by improving the LeGO-LOAM algorithm and dynamically adjusting the threshold settings, we effectively optimizes the point cloud segmentation and cleaning process, thereby enhancing localization accuracy and robustness. Compared to traditional fixed threshold settings, the dynamic threshold can better adapt to changes in different environments and scenes, making the algorithm more robust and stable. 

\section{Conclusion}
In this work, we propose an adaptive denoising-enhanced LiDAR odometry, greatly improving the robustness and resilience to degeneration in diverse terrains. This enhancement involves the precise selection of key feature points and the dynamic assessment of point cloud data reliability. The experimental validation, conducted on the KITTI benchmark and in real-world scenarios, demonstrates that our approach not only elevates the accuracy of the LeGO-LOAM system but also strengthens its robustness across different terrains.

We recognize that even widely adopted systems like LeGO-LOAM encounter difficulties with point cloud noise. The enhancements presented in this study effectively address these challenges, indicating a promising avenue for integration. Moving forward, our goal is to apply these improvements to a wider array of SLAM systems, delving into their adaptability and optimization potential within various SLAM frameworks. Through continuous research, we are committed to advancing SLAM technology for use in dynamic and variable environments, contributing to the evolution of autonomous navigation and mapping.

\bibliography{bare_jrnl_new_sample4}

\begin{IEEEbiography}[{\includegraphics[width=1in,height=1.25in,clip,keepaspectratio]{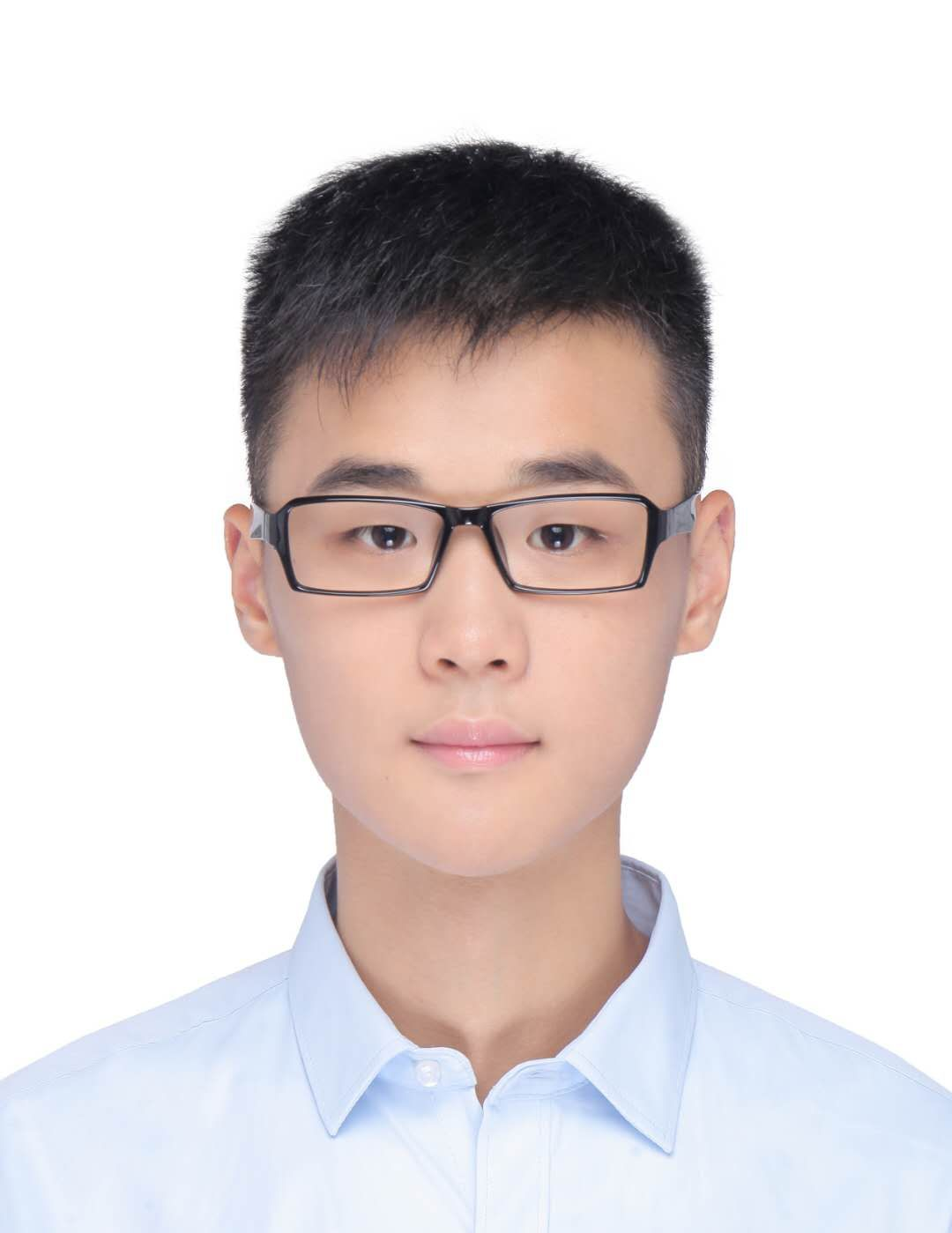}}]{Mazeyu Ji}
received the B.S. degree in control science and engineering from Tongji University, Shanghai, China, in 2023. He is currently pursuing a Master's degree in Electrical and Computer Engineering at the University of California, San Diego. His research interests include light detection and ranging (LiDAR) simultaneous localization and mapping (SLAM), point cloud processing and navigation of mobile robot.
\end{IEEEbiography}

\begin{IEEEbiography}[{\includegraphics[width=1in,height=1.25in,clip,keepaspectratio]{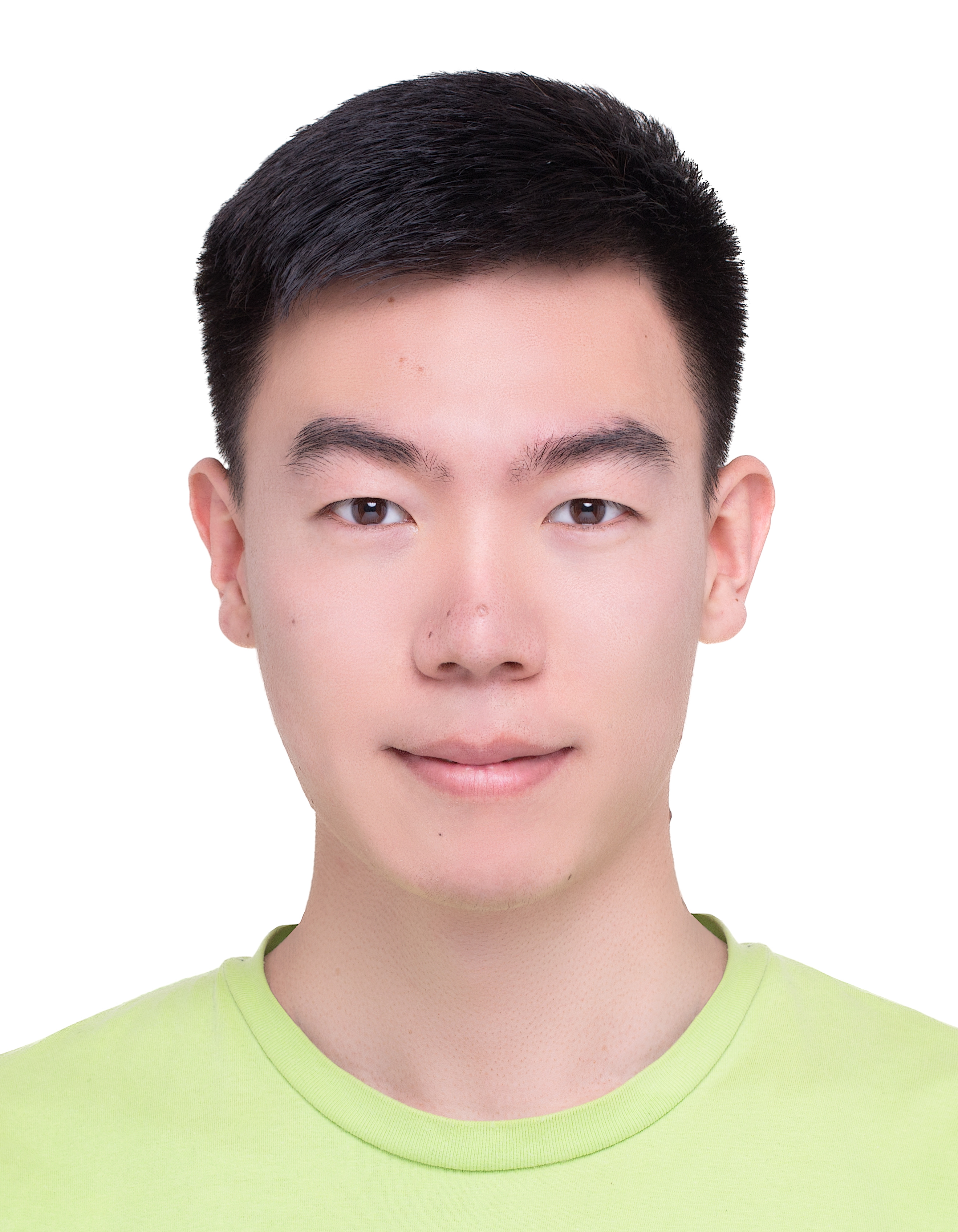}}]{Wenbo Shi}
received the B.S. degree in control science and engineering from 
Tongji University, Shanghai, China, in 2018. He is currently pursuing the 
Ph.D. degree in control science and engineering with Tongji University, 
Shanghai, China.
His research interests include light detection and ranging (LiDAR) 
simultaneous localization and mapping (SLAM), state estimation 
and navigation of mobile robot. 
\end{IEEEbiography}

\begin{IEEEbiography}[{\includegraphics[width=1in,height=1.25in,clip,keepaspectratio]{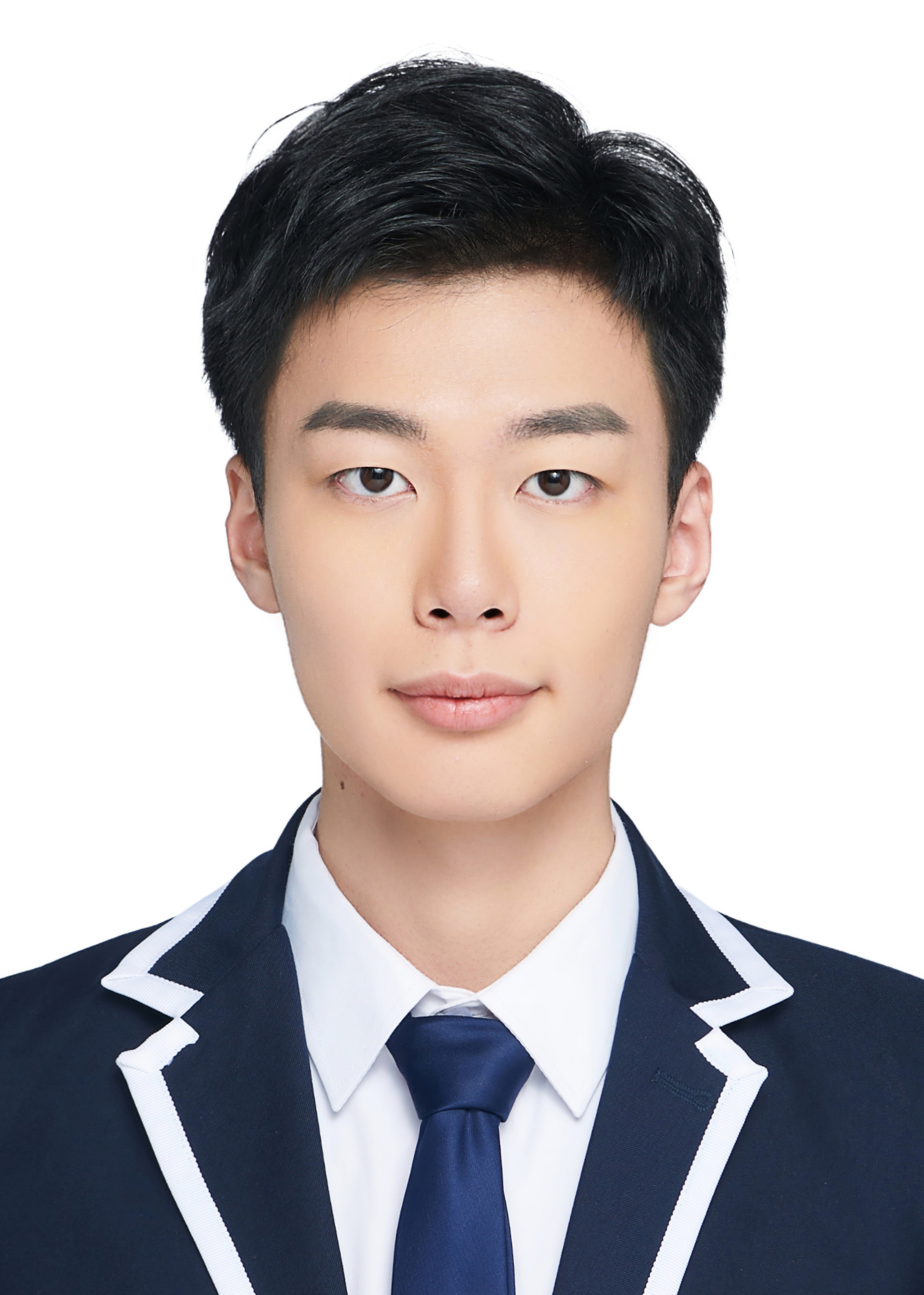}}]{Yujie Cui}
is currently pursuing the B.S. degree in control science and engineering with Tongji University, Shanghai, China. His research interests include light detection and ranging (LiDAR) simultaneous localization and mapping (SLAM).
\end{IEEEbiography}

\begin{IEEEbiography}[{\includegraphics[width=1in,height=1.25in,clip,keepaspectratio]{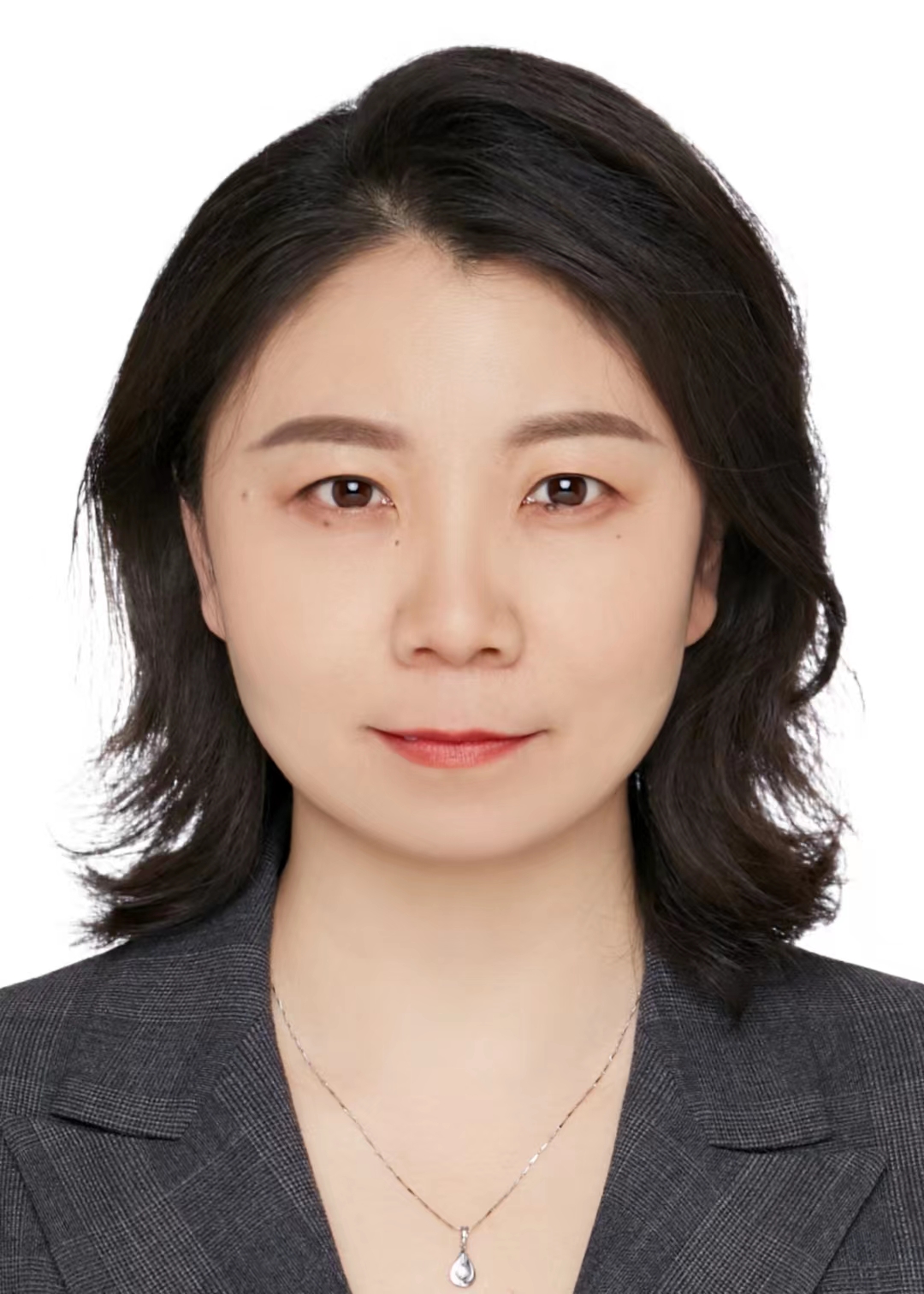}}]{Chengju Liu} 
received a Ph.D. in Control Theory and Control Engineering from Tongji 
University, Shanghai, China, in 2011. From October 2011 to July 2012, she 
worked in the BEACON Centre at Michigan State University, East Lansing, 
USA, as a research associate. From March 2011 to June 2013, she was a 
postdoctoral researcher, and she is currently a professor at the College 
of Electrical and Information Engineering, Tongji University, Shanghai, 
China.

Her research interests include intelligent control, motion control of 
legged robots, and evolutionary computation.
\end{IEEEbiography}

\begin{IEEEbiography}[{\includegraphics[width=1in,height=1.25in,clip,keepaspectratio]{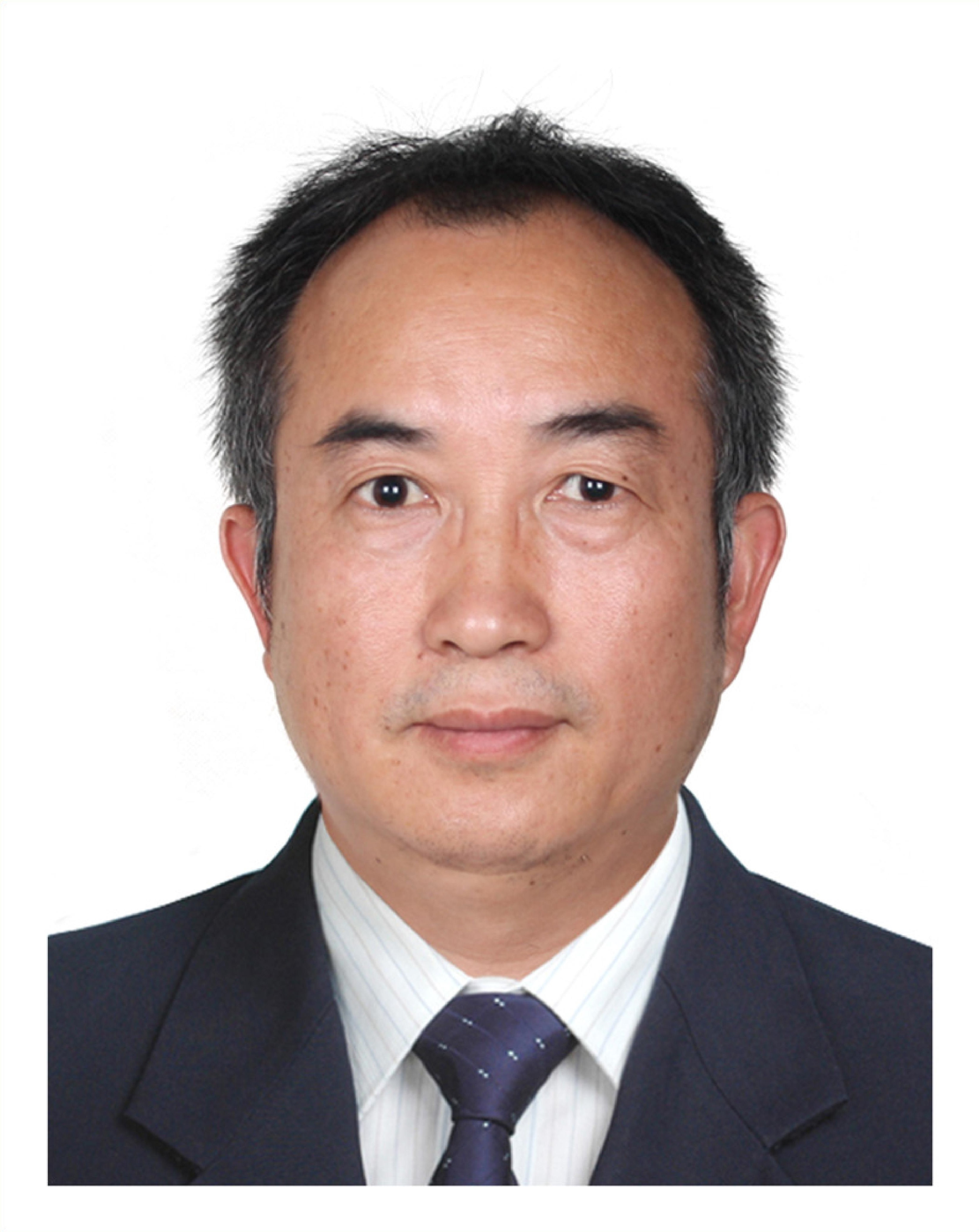}}]{Qijun Chen} 
received a B.S. in Automatic Control from Huazhong University of Science 
and Technology, Wuhan, China, in 1987, an M.S. in Control Engineering 
from Xi'an Jiaotong University, Xi'an China, in 1990 and a Ph.D. in 
Control Theory and Control Engineering from Tongji University, Shanghai, 
China, in 1999. He was a visiting professor at the University of 
California, Berkeley, CA, USA, in 2008. He is currently a professor in 
the College of Electronic and Information Engineering, Tongji University, 
Shanghai, China. 

His current research interests include network-based control systems and 
robotics.
\end{IEEEbiography}

\vfill

\end{document}